%% file: main_cr.tex
\documentclass{article}

\usepackage[nonatbib,final]{neurips_2025}
\usepackage[square,numbers,sort&compress,numbers]{natbib}

\usepackage[utf8]{inputenc} %
\usepackage[T1]{fontenc}    %
\usepackage{hyperref}       %
\usepackage{url}            %
\usepackage{booktabs}       %
\usepackage{amsfonts}       %
\usepackage{nicefrac}       %
\usepackage{microtype}      %
\usepackage{multirow}

\hypersetup{
    colorlinks,
    linkcolor={red!80!black},
    citecolor={blue!80!black},
    urlcolor={blue!80!black}
}

\title{Increasing the Utility of Synthetic Images\\through Chamfer Guidance}

\author{%
    Nicola Dall'Asen$^{1,2}$\thanks{Work done during internship at FAIR, Meta. Currently at Fondazione Bruno Kessler. $^{\dagger}$ equal contribution.} \And Xiaofeng Zhang$^{3,4,5\dagger}$ \And Reyhane Askari-Hemmat$^{4\dagger}$ \And Melissa Hall$^{4}$ \And Jakob Verbeek$^{4}$ \quad Adriana Romero-Soriano$^{3,4,6,7}$ \quad Michal Drozdzal$^{4}$\\
    \\
    $^1$University of Trento \quad
    $^2$University of Pisa \quad
    $^3$Mila - Qu\'{e}bec AI Institute \quad
    $^4$FAIR at Meta \quad \\
    $^5$Universit\'{e} de Montr\'{e}al \quad
    $^6$McGill University \quad
    $^7$Canada CIFAR AI chair\\
}
\input{preamble}

\begin{document}

\maketitle

\input{sec/0_abstract}

\input{sec/1_intro}

\input{sec/2_related}

\input{sec/3_method}

\input{sec/4_experiments}

\input{sec/5_conclusions}

\clearpage
{
\small
\bibliographystyle{plainnat}
\bibliography{main}
}

\input{sec/10_appendix}

\end{document}

%% file: preamble.tex
\usepackage{algorithm}
\usepackage{algpseudocode}
\usepackage{multirow}
\usepackage[dvipsnames,table,xcdraw]{xcolor}
\usepackage{amsmath} 
\usepackage{graphicx}
\usepackage{siunitx}     %
\usepackage{makecell}    %
\usepackage{pifont}%
\usepackage[normalem]{ulem}
\usepackage{enumitem}
\usepackage{subcaption}
\usepackage{xspace}

\newcommand{\xmark}{\ding{55}}%

\newcommand{\eg}{\textit{e.g.},~}
\newcommand{\wrt}{\textit{w.r.t.}~}
\newcommand{\ie}{\textit{i.e.},~}

\newcommand{\pp}[1]{\noindent\textbf{#1}}

\newcommand{\ldmonefour}{LDM{\textsubscript{1.4}}\xspace}
\newcommand{\ldmtwoone}{LDM{\textsubscript{2.1}}\xspace}
\newcommand{\ldmxl}{LDM{\textsubscript{XL}}\xspace}
\newcommand{\ldmonefive}{LDM{\textsubscript{1.5}}\xspace}
\newcommand{\ldmthreefive}{LDM{\textsubscript{3.5M}}\xspace}
\newcommand{\objreg}{\texttt{\{object\} in \{region\}}~}
\newcommand{\synreg}{\texttt{\{synonym\} in \{region\}}~}

\usepackage{cleveref}
\crefname{section}{Sec.}{Secs.}
\crefname{table}{Tab.}{Tabs.}
\crefname{figure}{Fig.}{Figs.}
\Crefname{section}{Section}{Sections}
\Crefname{table}{Table}{Tables}
\Crefname{figure}{Figure}{Figures}

\newcommand{\recomputed}[1]{\textcolor{black}{#1}}

\newcommand{\highlight}[2][yellow]{\mathchoice%
  {\colorbox{#1}{$\displaystyle#2$}}%
  {\colorbox{#1}{$\textstyle#2$}}%
  {\colorbox{#1}{$\scriptstyle#2$}}%
  {\colorbox{#1}{$\scriptscriptstyle#2$}}}%

%% file: sec/0_abstract.tex
\begin{abstract}
Conditional image generative models hold considerable promise to produce infinite amounts of synthetic training data. 
Yet, recent progress in generation quality has come at the expense of generation diversity, limiting the utility of these models as a source of synthetic training data.
Although guidance-based approaches have been introduced to improve the utility of generated data by focusing on quality or diversity,
the (implicit or explicit) utility functions oftentimes disregard the potential distribution shift between synthetic and real data. In this work, we introduce \emph{Chamfer Guidance}: a training-free guidance approach which leverages a handful of real exemplar  images to characterize the quality and diversity of synthetic data. We show that by leveraging the proposed Chamfer Guidance, we can boost the diversity of the generations \wrt a dataset of real images while maintaining or improving the generation quality on ImageNet-1k and standard geo-diversity benchmarks. Our approach achieves state-of-the-art few-shot performance with as little as 2 exemplar real images, obtaining 96.4\% in terms of precision, and 86.4\% in terms of distributional coverage, which increase to 97.5\% and 92.7\%, respectively, when using 32 real images. We showcase the benefits of the Chamfer Guidance generation by training downstream image classifiers on synthetic data, achieving accuracy boost of up to 15\% for in-distribution over the baselines, and up to 16\% in out-of-distribution.
Furthermore, our approach does not require using the unconditional model, and thus obtains a 31\% reduction in FLOPs \wrt classifier-free-guidance-based approaches at sampling time.\looseness-1

\end{abstract}

%% file: sec/1_intro.tex
\section{Introduction}
\label{sec:intro}

\begin{figure*}[!th]
    \centering
    \includegraphics[width=1.0\linewidth]{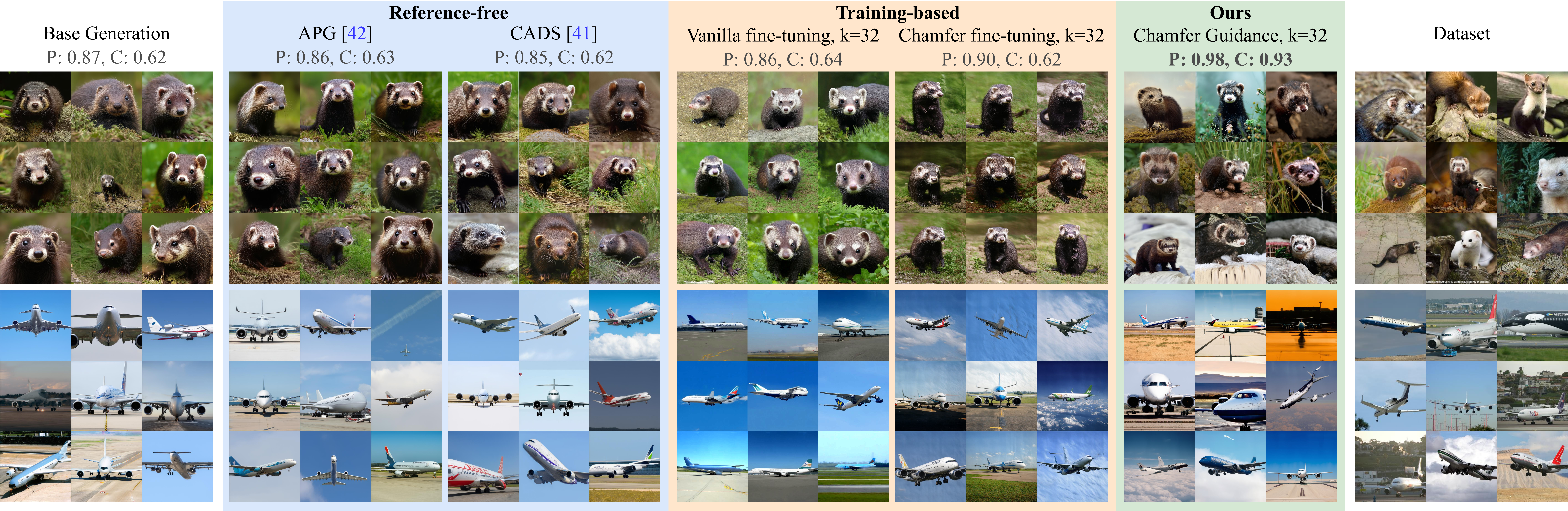}
    \caption{Our Chamfer Guidance addresses key limitations of existing image generation approaches, producing \textbf{high-quality} and \textbf{diverse} outputs. 
    Base models (here \ldmthreefive) necessitate high CFG scales to achieve prompt adherence and quality, at the expense of diversity. Reference-free methods can introduce \textit{ungrounded} diversity, failing to capture the underlying data distribution. While training-based solutions effectively narrow the fidelity gap with the reference distribution, they suffer from low subject diversity, particularly in backgrounds. Our Chamfer Guidance achieves superior image quality without using CFG, substantially improving grounded coverage (C) and aligning the generated images more precisely (P) with the reference distribution. Best viewed zoomed in.}
    \vspace{-2em}
    \label{fig:showcase}
\end{figure*}

In the last few years, conditional image generative models~\citep{sohl2015deep,ho2020denoising,rombach2021high} have demonstrated extraordinary capabilities, producing highly realistic images from both textual descriptions and class labels. These models have rapidly evolved from experimental research tools to widely accessible applications, enabling creative expression for users across various domains. With this acceleration, researchers have also started exploring the use-cases of conditional image generative models as synthetic training data for downstream machine learning models~\citep{azizi2023synthetic}. \looseness-1 %
However, recent research has revealed that as models grow in size and capability, they tend to produce images of higher quality, aligned with human preference, but with diminished diversity, hindering their utility as synthetic training data generators~\citep{astolfi2024pareto}.\looseness-1 %

Learning-based approaches and guidance-based sampling techniques~\citep{hemmat2023feedbackguided,askari25dp,hemmat2024improving} have been introduced in the literature to mitigate the shortcomings of vanilla synthetic data. On the one hand, learning-based approaches~\citep{gal2022imageworthwordpersonalizing,ruiz2023dreambooth} aim to improve the overall utility of synthetic data by reducing the distribution shift between generated samples and real samples. These techniques leverage real data samples to train or fine-tune parts of existing conditional image generative models. On the other hand, guidance-based sampling methods define reward functions to guide the process of generating synthetic data. These reward functions characterize the utility of the generated samples in terms of generation quality~\citep{xu2023imagereward} and diversity~\citep{hemmat2024improving}. \looseness-1 %
However, in-depth analyses of conditional image generative models have highlighted a tension between these two desirable properties~\cite{astolfi2024pareto}. This tension becomes more apparent when defining generation diversity as the variation among generated samples without reference to any target distribution. Contextualizing the desired diversity through real exemplar data holds the promise of limiting potentially disadvantageous variations in synthetic data.\looseness-1 %

Using exemplar data at inference time to guide the generation process is reminiscent of few-shot in-context learning (ICL) approaches in large language models (LLMs), where a small number of examples are used to adapt to new tasks without requiring model fine-tuning. Yet, in conditional image generation, most few-shot approaches require partial model training or fine-tuning~\citep{gal2022imageworthwordpersonalizing,ruiz2023dreambooth}, and the use of exemplar data \emph{at inference time} remains underexplored. Following this line of research, Contextualized Vendi Score Guidance (c-VSG)~\citep{hemmat2024improving} proposed to leverage a handful of contextualizing real images to ground the diversity of generations to that of the real data. However, contrary to ICL approaches~\citep{agarwal2024many}, increasing the number of examples did not show to improve c-VSG's performance~\citep{hemmat2024improving}.\looseness-1

In this work, we introduce a new training-free guidance method to increase the utility of synthetic data. The method leverages exemplar real data to characterize the desired quality and diversity of synthetic data and scales gracefully with the number of exemplar real images. We define utility of synthetic data through the Chamfer distance~\citep{wu2021balanced} between generated and real data, and use this utility formulation as reward guidance to improve off-the-shelf state-of-the-art conditional image generative models. 
We demonstrate that our Chamfer Guidance approach increases the diversity of generated images compared to previous methods, while maintaining or improving generation quality. We validate these results on ImageNet-1k~\citep{deng2009imagenet} and standard geo-diversity benchmarks~\citep{ramaswamy2023geode,gaviria2022dollar}.
On ImageNet-1k, we reach a distributional coverage of 91.2\% and 92.7\% while reaching a precision of 95\% and 97.5\% for \ldmonefive~\citep{rombach2021high} and \ldmthreefive~\citep{esser2024scaling}, respectively. To further illustrate the benefits of our approach, we contrast Chamfer Guidance with alternative approaches in Figure~\ref{fig:showcase}. On the geodiversity task, we observe an improvement \wrt prior art of 6.5\% and 5.7\% in terms of average $F_1$ and worst-region $F_1$, respectively. 
To validate the utility of our Chamfer Guidance, we train a classifier for the ImageNet-1k~\cite{deng2009imagenet} dataset on synthetic data, and we demonstrate that our guidance can boost downstream accuracy of up to 15\% over the classifier-free guidance sampling. We further investigate out-of-distribution generalization, and our guidance outperforms the classifier-free guidance sampling on several ImageNet variants, and it can obtain a boost of up to 16\% on ImageNet-Sketch~\cite{wang2019learning} using \ldmthreefive.
Finally, for \ldmthreefive, our approach does not require unconditional model to obtain state-of-the-art quality and diversity, obtaining a 31\% reduction in FLOPs \wrt CFG-based approaches during sampling. With this work, we advance the synthetic training data field and show that by designing appropriate guidance functions we can unlock the full potential of conditional image generative models and better exploit the knowledge encapsulated in them.\looseness-1

%% file: sec/2_related.tex
\section{Related work}
\label{sec:related}

While most modern text-to-image diffusion models utilize classifier-free guidance~\citep{ho2022classifier}, there has been increasing exploration of alternative guidance methods to further improve the sampling process.
A subset of works focus on improving image quality with methods including modifications to classifier-free guidance to reduce over-saturation, such as Adaptive Projected Guidance (APG)~\citep{sadat2024eliminatingoversaturationartifactshigh}, utilization of blurring~\citep{hong2023improving,hong2024smoothed} or perturbations~\citep{ahn2024self}, and employment of a smaller, less-trained version of the model for guidance~\citep{karras2024guiding}.
Other works explore guidance methods to improve the diversity of generated images,  \eg~Condition-Annealed Diffusion Sampler (CADS)~\cite{sadat2023cads} anneals the conditioning signal by incorporating scheduled, monotonically decreasing Gaussian noise into the conditioning vector during the diffusion chain to increase diversity of the generated samples, while Limited Interval guidance~\citep{kynkaanniemi2024applying} removes the conditioning signal in a certain interval to improve ungrounded diversity. Particle Guidance~\citep{corso2024particle} overcomes the common assumption of independent samples in a batched generation, and proposes an extension of sampling where a joint-particle time-evolving potential enforces diversity.
c-VSG, which guides the denoising process to increase the diversity of a sample compared to images previously generated with the same prompt~\citep{hemmat2024improving}.
Still other works focus on safety-related guidance by \textit{reducing} harmful representations, such as using classifier guidance to guide generations away from inappropriate content~\cite{brack2023mitigating,schramowski2023safe} or ``forgetting'' concepts by zeroing out their cross-attention scores~\citep{zhang2023forget}. 
However, these methods tend to focus on generation quality/diversity/safety as ends in themselves, rather than as means to increasing a synthetic image corpus' \textit{utility} in training downstream models.

Other recent works have begun to explore use of guidance methods to augment real image training datasets with synthetic samples for use in downstream tasks, such as using the loss and entropy of the downstream classifer to guide the synthetic image generator~\citep{hemmat2023feedbackguided}, promoting sample diversity to supplement small-scale datasets~\citep{zhang2023expanding}, and leveraging textual inversion ~\citep{gal2022imageworthwordpersonalizing} to generate images for the use of long-tail image classification~\citep{shin2023fill}.
There has also been increasing study into the use of guidance to help in model self-improvement, such as using generated samples as negative guidance to avoid mode collapse/degradation~\citep{alemohammad2024self}.
Our work complements these methods by utilizing features from a small set of real images in the guidance process to improve the utility of generated samples. 

Finally, in lieu of guidance approaches to improve the generations of diffusion models, a standard practice is to perform fine-tuning.  
These methods include fine-tuning directly on additional data sources that better capture the desired qualities of the generations~\citep{esser2024scaling,rombach2021high,dai2023emu,esposito2023mitigating} or utilizing reward models to incorporate preferences of model outputs~\citep{xu2023imagereward,dong2023raft,clark2024directlyfinetuningdiffusionmodels}. For example, it is possible to directly fine-tune latent diffusion models by integrating human-preference feedback during the denoising process using Reward Feedback Learning (ReFL)~\cite{xu2023imagereward}. 
Instead of the conventional generate-then-filter approach, ReFL leverages a reward model to provide gradient feedback. A key insight is that the quality of generated images becomes discernible only in the later denoising steps. At early timesteps, the reward scores are uniformly low, but once a sufficient number of steps have been taken (\eg, after 30 out of 40 steps), the reward model produces distinguishable scores. To balance stability and effective feedback, a random timestep from a designated (late) interval is chosen for fine-tuning. Although this approach has been proven successful for quality fine-tuning, it is not yet tested whether this approach would increase diversity in generation given the proper reward. 
In this work, we study how strong diversity and quality can be achieved by guidance \textit{without} the need to perform fine-tuning.

%% file: sec/3_method.tex
\section{Method}
\label{sec:method}

In this section, we introduce a training-free approach for diffusion and flow models, which leverages real data samples to characterize the desired quality and diversity of the synthetic data. %
Our method leverages the Chamfer distance as a mechanism to match sets of real and generated samples.\looseness-1 %

\subsection{Preliminaries}
\label{sec:prelim}

\pp{Diffusion Models.}
Diffusion models~\cite{sohl2015deep,ho2020denoising} are generative models that transform a unit Gaussian prior into a data distribution via iterative denoising. They consist of a forward and a backward process. 
The forward process progressively adds noise to data $x_0$ over $T$ timesteps following a Markovian process $q(x_t | x_{t-1}) = \mathcal{N}(x_t; \sqrt{1 - \beta_t} x_{t-1}, \beta_t \mathbf{I})$, where $\beta_t \in (0, 1)$ is a variance schedule.
The backward process learns to reverse this by denoising a sample $x_t$ to $x_{t-1}$, modeled as $p_\theta(x_{t-1} | x_t) = \mathcal{N}(x_{t-1}; \mu_\theta(x_t, t), \Sigma_\theta(x_t, t))$. 
\citet{ho2020denoising} show how we can equivalently train a denoising neural network $\epsilon_\theta(x_t, t)$ to predict the noise $\epsilon$ using the following objective:
\begin{equation}
\mathcal{L}_\text{simple} = \mathbb{E}_{t, x_0, \epsilon} \left[ \left\| \epsilon - \epsilon_\theta\left(\sqrt{\bar{\alpha}_t} x_0 + \sqrt{1 - \bar{\alpha}_t} \epsilon, t\right) \right\|^2 \right]. \label{eq:objective}
\end{equation}
Sampling starts with $x_T \sim \mathcal{N}(0, \mathbf{I})$, and the learned reverse process is iteratively applied to obtain $x_0$, \ie $x_{t-1} = \text{denoise}(x_t, \epsilon_{\theta}) := \frac{1}{\sqrt{\alpha_t}} \left(x_t - \frac{1-\alpha_t}{\sqrt{1-\bar{\alpha}_t}} \epsilon_\theta(x_t, t)\right) + \sigma_t\mathbf{\varepsilon}$, with $\varepsilon~\sim \mathcal{N}(0, \mathbf{I})$, and $\sigma$ a weighting parameter~\cite{ho2020denoising}.\looseness-1

\pp{Flow matching.} Flow Matching (FM) is a generative modeling approach that offers a deterministic alternative to diffusion models by learning continuous transformations between probability distributions. While diffusion models rely on stochastic processes that gradually add and then remove noise through stochastic differential equation (SDE), FM employs first-order ordinary differential equations (ODEs) to define a vector field guiding the transformation from a simple base distribution to the target data distribution. This deterministic nature results in faster inference and reduced computational costs. Moreover, FM can work with optimal transport interpolations, which can lead to more direct and stable mappings compared to the often curved trajectories in diffusion models~\cite{lipman2023flow}.

\pp{CFG.} Classifier-free guidance (CFG)~\cite{ho2022classifier} enhances controllability in diffusion models by combining conditional and unconditional outputs during sampling as follows:
\begin{equation}
    \label{eq:cfg}
    \epsilon_{\text{guidance}}(x_t, t, c) = (1 + \omega)\epsilon_{\theta}(x_t, t, c) - \omega\epsilon_{\theta}(x_t, t),
\end{equation}
where \(\omega\) adjusts guidance strength, \(\epsilon_{\theta}(x_t, t, c)\) is the noise prediction conditioned on  $c$ (commonly a class label or  text prompt), and \(\epsilon_{\theta}(x_t, t)\) is the unconditional counterpart. This extrapolation trades diversity for fidelity without requiring auxiliary  classifiers.\looseness-1

\pp{Classifier-guidance.} Another approach to control the generation is classifier-guidance \citep{dhariwal2021diffusion}, which uses a pretrained classifier $p_\phi(y|x_t)$ to modify the unconditional score function $\nabla_{x_t} \log p_\theta(x_t)$, yielding the following modified score function for conditional generation:
\begin{equation}
\nabla_{x_t} \log p_{\theta}(x_t|y) = \nabla_{x_t} \log p_\theta(x_t) + \gamma \nabla_{x_t} \log p_\phi(y|x_t),
\label{eq:classifier_guidance}
\end{equation}
where $\gamma$ is a scaling factor that controls the strength of the guidance signal. Classifier guidance can also be used on top of the CFG score prediction from \Cref{eq:cfg}.

\pp{DDIM.} At sampling time, using the insights of DDIM~\cite{song2020denoising}, we can obtain an approximate denoised image for each timestep $t$,
\begin{equation}\label{eq:DDIM_approx}
    \hat{x}_{0, t} = \text{DDIMApprox} (x_t) := \frac{x_t - \sqrt{1 - {\bar{\alpha}}_t} \epsilon_\theta(x_t)}{\sqrt{{\bar{\alpha}}_t}}.
\end{equation}

\noindent Access to the approximated denoised sample is useful and necessary to apply guidance on the denoising process at inference time without modifying the diffusion model network $\epsilon_{\theta}$~\cite{hemmat2024improving}.

\subsection{Chamfer distance for image distribution matching}

The Chamfer distance~\citep{barrow1977parametric} is a metric used to quantify the similarity between two sets of points by measuring the average closest-point distance between each point in one set to the other and vice versa. It has been widely used in 3D computer vision and computer graphics  to register point clouds~\citep{sun2018pix3d}.\looseness-1 %

In this work, we represent image distributions as sets of points and leverage the Chamfer distance for distribution matching. Given a set of real images $\mathcal{X}$ and a set of generated images $\mathcal{Y}$, our goal is to encourage the generated set of images to be as close as possible to a set of real target images. We compute the Chamfer distance between these sets of images as: 
\begin{equation}\label{eq:chamfer_loss}
\resizebox{.9\hsize}{!}{
$\mathcal{L}_{\text{Chamfer}}(\mathcal{X}, \mathcal{Y}) = \highlight[NavyBlue!17]{\frac{1}{|\mathcal{X}|}\sum_{x \in \mathcal{X}}\min_{y \in \mathcal{Y}}||x-y||^2} + \highlight[ForestGreen!17]{\frac{1}{|\mathcal{Y}|}\sum_{y \in \mathcal{Y}}\min_{x \in \mathcal{X}}||x-y||^2}$
}.
\end{equation}

The distance consists of two terms: %
The {\textcolor{NavyBlue}{first term}} matches each real sample with its closest generated counterpart. %
This term is reminiscent to  {\textcolor{NavyBlue}{\emph{diversity}}} metrics used to evaluate synthetic data~\citep{kynkaanniemi2019improved,naeem2020reliable}, and may be viewed as a particular instance of Implicit Maximum Likelihood Estimation~\cite{li2018implicit}, which inherently resists mode collapse by encouraging coverage of the entire data distribution. 
The {\textcolor{ForestGreen}{second term}} matches each generated sample with its closest real counterpart. %
This term is reminiscent to  {\textcolor{ForestGreen}{\emph{fidelity}}} metrics used to evaluate synthetic data~\citep{kynkaanniemi2019improved,naeem2020reliable}.

\pp{Representation space projection.}
Before computing the Chamfer distance, we project both real and synthetic images into a representation space that induces more semantically meaningful distances than a naive  $\ell_2$ metric in RGB pixel-space.
 We use DINOv2~\cite{oquab2023dinov2}, which captures semantic information through self-supervised learning. Unlike CLIP's contrastive learning with text-image pairs and Inception's supervised ImageNet training, DINOv2's self-supervised learning approach provides a feature representation that  balances between focusing on important objects and holistic image structure~\cite{stein2023exposing}. 
 Moreover, DINOv2's features have been shown to capture human-perceived similarity better~\citep{hall2024towards} than representations such as CLIP~\cite{radford2021learning} and Inceptionv3~\citep{szegedy2016rethinking}.\looseness-1

\subsection{Inference-time guidance}
To improve the alignment between the generated and real data distributions at inference time, without requiring any fine-tuning, we introduce a guidance mechanism based on the Chamfer distance. This builds on the general framework of guidance in diffusion models, where external signals are used to steer the generative process toward a desired objective \citep{dhariwal2021diffusion, sehwag2022generating, hemmat2023feedbackguided, askari25dp, hemmat2024improving}. 

\pp{Chamfer Guidance.}
We introduce Chamfer Guidance, where the external signal is derived from the Chamfer distance between a batch of generated samples and a batch of real samples. Let $\mathcal{X}$ denote a set of features extracted from real images, and let $\hat{x}_{0,t}$ be the denoised approximation of a batch of generated samples $x_t$ at time step $t$  (see \Cref{eq:DDIM_approx}). The guidance score becomes:
\begin{equation}
\nabla_{x_t} \log p_{\theta}(x_t|c, \mathcal{X}) = \nabla_{x_t} \log p_\theta(x_t|c) - \gamma \nabla_{x_t} \mathcal{L}_{\text{Chamfer}}(\mathcal{X}, \hat{x}_{0,t}),
\label{eq:chamfer_guidance}
\end{equation}
where $\mathcal{L}_{\text{Chamfer}}$ is the Chamfer distance between the real sample set $\mathcal{X}$ and the current batch of denoised samples, as defined in~\Cref{eq:chamfer_loss}. The negative sign in \Cref{eq:chamfer_guidance} reflects the goal of minimizing the Chamfer distance, thereby encouraging the generated samples to be close to the real data (fidelity) and to cover its modes (diversity). 

\pp{Efficient approximation.} Evaluating the Chamfer distance across the full reverse diffusion trajectory is computationally expensive. To mitigate this, similar to \citep{hemmat2023feedbackguided, askari25dp}, we adopt the DDIM approximation in \Cref{eq:DDIM_approx}, which provides a first-order estimate of the final denoised sample at each time step $t$. 
This allows us to compute a differentiable reward signal and its gradient during the intermediate steps of the diffusion process without completing all $T$ reverse steps. 

\pp{Chamfer as a reward signal.}
Chamfer Guidance can also be interpreted in the context of reward-based generation methods, such as ReFL. In our experiments we find, however, that applying Chamfer-based guidance at inference time leads to stronger distributional alignment than using the same reward function for fine-tuning.

%% file: sec/4_experiments.tex
\section{Experiments}
\label{sec:exp}
{

\subsection{Experimental setup} 
We consider three primary experimental settings. The first is an \textit{object-centric} setup, where the focus lies on guiding or evaluating the generation of specific object categories. This setting allows us to assess the capability of conditional diffusion models in producing accurate and semantically consistent visual representations of well-defined object classes. The second setting addresses \textit{geographic representation} and targets the well-documented limitations of T2I models in handling geographically grounded content. In this setting, we aim to evaluate how well the models capture regional diversity and mitigate the visually biased depiction of locations~\cite{hall2023dig,hemmat2024improving}. Finally, we evaluate the downstream utility of the data generated by our Chamfer Guidance by training an image classifier purely on synthetic data.\looseness-1 

\pp{Datasets.} We utilize three publicly available datasets. For the object-centric setting, we use ImageNet-1k~\cite{deng2009imagenet}, a large-scale image classification dataset containing over one million images across 1,000 object categories. %
For the geodiversity representation, we use GeoDE~\cite{ramaswamy2023geode} and DollarStreet~\cite{gaviria2022dollar}. 
GeoDE contains a curated set of images annotated with geographic provenance and object labels. %
DollarStreet includes photographs of everyday household items from around the world, labeled by country and income level, thus enabling fine-grained analysis of regional visual representations, in line with prior works~\cite{hall2023dig,hall2024towards,hemmat2024improving}. For downstream utility, we employ ImageNet-1k~\cite{deng2009imagenet}. We further employ ImageNet-V2~\cite{recht2019imagenet}, ImageNet-Sketch~\cite{wang2019learning}, ImageNet-R~\cite{hendrycks2021many}, and ImageNet-A~\cite{hendrycks2021nae} to measure out-of-distribution generalization.

\pp{Models.} We consider two state-of-the-art T2I models, in particular latent diffusion models (LDMs), \ldmonefive~\cite{rombach2021high}, and  \ldmthreefive
~\cite{esser2024scaling}, which we use in a class-conditional and text-conditional way.\looseness-1

\pp{Implementation Details.} All experiments are implemented using the \texttt{diffusers} library~\cite{von-platen-etal-2022-diffusers}, using the default samplers with 40 denoising steps. For the latent projection of our Chamfer Guidance, we primarily use the DINOv2~\cite{oquab2023dinov2} (ViT-L) feature space, which offers strong semantic representations suited for fine-grained perceptual alignment and whose features have been shown to correlate better with human judgement of similarity~\citep{hall2024towards}. To enable comparison with prior studies~\cite{hemmat2024improving}, we also report results using CLIP embeddings~\cite{radford2021learning} for the geographic diversity scenario. Evaluation in the object-centric setting is conducted using the \texttt{dgm-eval} library~\cite{stein2023exposing}, while the geographic representation scenario utilizes the publicly available evaluation code from DIG-In~\cite{hall2023dig}. The Chamfer distance implementation is from \texttt{PyTorch3D} library~\cite{ravi2020pytorch3d}. As in c-VSG~\cite{hemmat2024improving}, we set the inference-time guidance frequency to $G_{\text{freq}} = 5$, \ie we apply  Chamfer Guidance  once every five denoising steps. 
}

\subsection{Object-centric scenario: ImageNet-1k}

\pp{Baselines.} We employ reference-free, and reference-based baselines, where reference-based baselines use a few real data samples as reference.
As reference-free baselines, we test the default capabilities of LDMs with different CFG~\cite{ho2022classifier} values, as well as advanced guidance techniques such as APG~\citep{sadat2024eliminatingoversaturationartifactshigh}, CADS~\cite{sadat2023cads}, Limited Interval guidance~\citep{kynkaanniemi2024applying}, and Particle Guidance~\citep{corso2024particle}.  %
Regarding reference-based approaches, for \ldmonefive we test c-VSG~\cite{hemmat2024improving}, which employs a memory bank of prior outputs and real-world exemplar images to guide the generation process, balancing ungrounded and grounded diversity through the use of two Vendi Scores~\cite{friedman2022vendi}. 
Additionaly, we employ two training-based solutions, vanilla fine-tuning and reward-based fine-tuning~\cite{xu2023imagereward}. For vanilla fine-tuning, we fine-tune the model through the standard denoising loss in \Cref{eq:objective}, using the set of real reference images $\mathcal{X}$ as fine-tuning data. Drawing inspiration from ImageReward~\cite{xu2023imagereward} and their ReFL algorithm, we implement a reward-based fine-tuning approach which leverages a reward function derived from the negative Chamfer distance, \ie $r = -\mathcal{L}_{\text{Chamfer}}$. As such, this approach requires a few real data samples to encourage the model to generate samples that minimize the distributional distance to the real data. Unlike traditional fine-tuning which focuses on individual samples, this approach explicitly optimizes for distribution-level properties.\looseness-1 

\input{tables/in1k_unified_omega}

{
\pp{Metrics.} To comprehensively assess both image quality and diversity aspects, we report Precision and Recall~\cite{kynkaanniemi2019improved}, Density and Coverage~\cite{naeem2020reliable}, $F_1$ measured as the harmonic mean of Precision to Coverage, as well as Fréchet Distance (FD)~\cite{heusel2017gans}. Note that these metrics are grounded on a reference dataset. We use DINOv2 features to compute all the metrics as recent literature~\cite{stein2023exposing,hall2024towards} found that DINOv2 space provides more accurate estimations of perceptual similarity. In addition, we report the standard FD using InceptionV3 features.\looseness-1

\pp{Number of real images.} We investigate the effect of increasing the number of real images used in both fine-tuning and inference-time guidance approaches. This allows us to establish the data efficiency and scaling properties of each method. 
For the  number of real images we consider $k \in [1,2,4,8,16,32]$.\looseness-1

\pp{Implementation details.} For all the experiments with $k \leq 8$, we use a single H100 GPU to perform training and inference. We use multiple GPUs for $k=[16,32]$. Vanilla fine-tuning and Chamfer fine-tuning solutions are trained for at most 5,000 steps with checkpoints taken every 1,000 steps. We use a constant learning rate of $10^{-6}$ across all experiments. For \ldmonefive, we fine-tune the entire U-Net backbone, while for \ldmthreefive we employ LoRA~\cite{hu2022lora} fine-tuning with a rank $r=4$ applied on the key, query, value, and output layers of attention modules. For ReFL hyperparameters, we use the official implementation and set $\lambda = 10^{-3}, T=40, T_1 = 30, T_2=39$.\looseness-1
}

\subsubsection{Quantitative Results}

We show the quantitative evaluation  on ImageNet-1k dataset for \ldmonefive~and \ldmthreefive in \Cref{tab:quantitative_sd15_in1k,tab:quantitative_sd35_in1k}, respectively. 
For each $k$, we report the metrics corresponding to the best $F_1$ score \wrt the validation set of each dataset. 
Our analysis reveals that existing reference-free solutions are unable to deliver satisfactory outcomes, offering only marginal improvements over the base LDM performance in terms of diversity. c-VSG achieves the best recall, at the expense of coverage, indicating the generation of outlier images.
Furthermore, we observe that neither vanilla fine-tuning nor Chamfer fine-tuning improve their performance when trained with an increased number of available training samples. Interestingly, on \ldmthreefive our Chamfer Guidance obtains state-of-the-art results in fidelity and diversity without using CFG, \ie $\omega=1.0$. This brings a significant computational reduction \wrt CFG-based approaches of $\approx31\%$ for the case $k=32$. The detailed computation of efficiency is presented in the Appendix. \\
\Cref{fig:scaling_imagenet_1k} shows how our Chamfer Guidance effectively increases the diversity when a higher number of real samples become available. Meanwhile, training-based approaches only marginally increase precision when learning on higher number of real samples $k$, while not improving the image quality as indicated by the FDD metric. Our results on Chamfer Guidance are in line with recent trends in LLMs, where test-time compute can be more effective than training or fine-tuning-time compute~\cite{snell2024scaling}.

\begin{figure*}[t]
    \centering
    \includegraphics[width=0.9\linewidth]{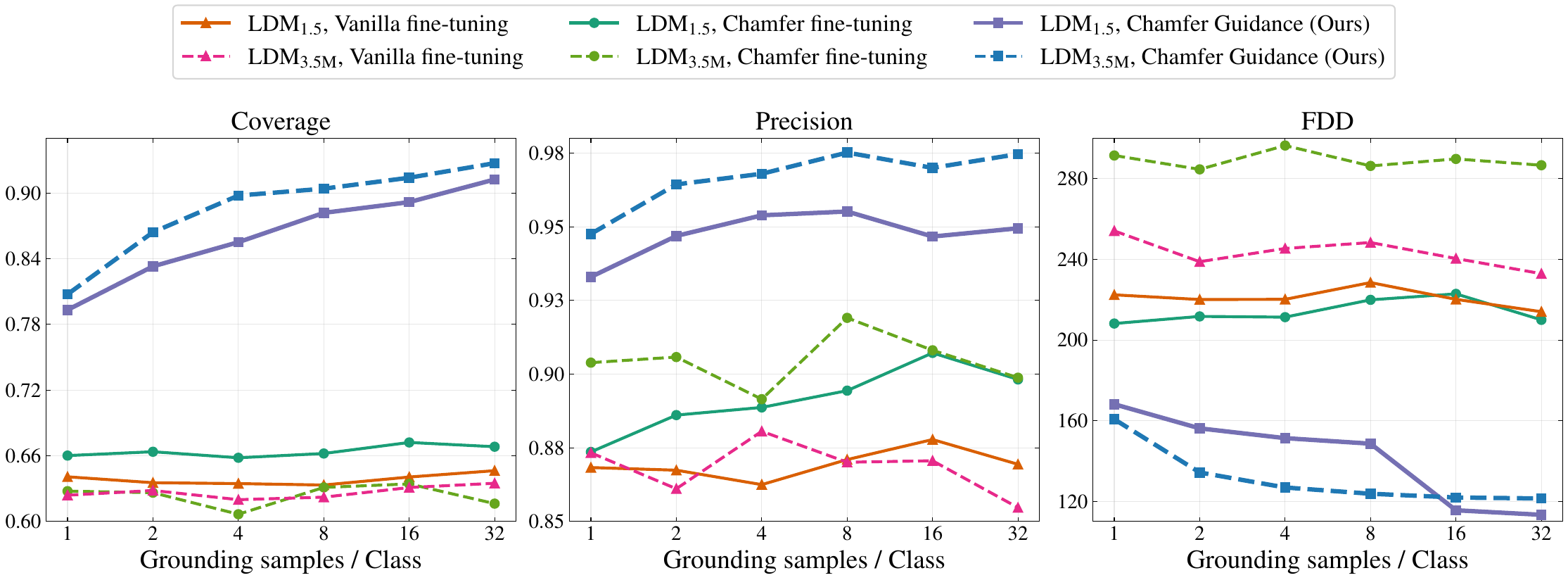}
    \caption{Effect of the number of  real reference samples $k$ on  \ldmonefive~ and \ldmthreefive~ for ImageNet-1k. We can see that only our Chamfer Guidance can effectively leverage the  the increased number of reference images, consistently obtaining  favorable  trends across  Coverage,  Precision, and  FDD.}
    \label{fig:scaling_imagenet_1k}
    \vspace{-2.em}
\end{figure*}

\subsection{Geographic Diversity}

We conduct a thorough comparative analysis of our approach against existing geographic diversity enhancement techniques, particularly focusing on the contextualized Vendi Score Guidance (c-VSG) method~\cite{hemmat2024improving}, which is the closest in spirit to our approach. This comparison reveals several key advantages of our Chamfer Guidance.\looseness-1\\
\pp{Datasets and evaluation metrics.} We follow the same evaluation protocol of c-VSG, but we employ the $F_1$ score between precision and coverage metrics instead of precision and recall. This is to avoid inflated results due to generated outliers. We use Inception-based metrics and prompt the models with \texttt{\{object\} in \{region\}} for a fair comparison with c-VSG. We report results computed in DINOv2 space in the Appendix. When reporting results for ``worst region'', we follow the original protocol to first identify the worst region in terms of the $F_1$ score, and then report the other metrics corresponding to the same region. 
This may lead to cases where the worst region has a higher value for the metric \wrt the average.\looseness-1\\
\pp{Implementation details.} We use \ldmonefive~as in c-VSG and apply the same filtering on the GeoDE dataset as described in prior work~\cite{hall2023dig,hemmat2024improving}.
We choose the DINOv2~\cite{oquab2023dinov2} latent space to compute the Chamfer distance, but we also report results using CLIP~\cite{radford2021learning} for a fair comparison with c-VSG.\looseness-1

\subsubsection{Quantitative Results}

\input{cvsg_tables/cvsg_geode_pc}

To enable a direct comparison, our experiments replicate the settings of c-VSG~\cite{hemmat2024improving}.~\Cref{tab:cvsg_geode_pc} presents a comparative analysis of our Chamfer Guidance against c-VSG on the GeoDE dataset, using recomputed baselines to use the revised $F_1$ score. Details about the baselines are presented in the Appendix. Notably, our optimal approach (DINOv2 as the feature extractor, and 4 grounding samples) achieves a nearly $7\%$ improvement in $F_1$  score over c-VSG. We also observe a consistent scaling in coverage, exhibiting the same trend as in the object-centric scenario. This indicates that the observed behavior persists even in the more challenging context of geographical diversity. Furthermore, our method substantially enhances worst-region coverage by up to $4.9\%$, demonstrating its efficacy in mitigating regional misrepresentation. As expected, the highest image-text alignment, measured by CLIPScore, is obtained when employing the CLIP latent space for projecting images during Chamfer distance computation. This methodology effectively guides synthetic images towards the CLIP subspace corresponding to the target object. Then, we evaluate our approach on DollarStreet, with results presented in the Appendix, where we demonstrate a $3.5\%$ improvement in the $F_1$ score compared to c-VSG, due to the enhanced quality and diversity of the generated data. \\
In summary, we can draw the following conclusions from these results: Chamfer Guidance enhances the quality (precision) and diversity (coverage) of generated data, and demonstrates superior image-text alignment. Unlike c-VSG, it exhibits favorable scaling with increased real data volume. We show qualitative results in the Appendix, where our Chamfer Guidance showcases increased quality, less saturation and more diversity.\\
Our analysis reveals two additional significant advantages of our approach: (1) Unlike c-VSG, which necessitates storing intermediate results to continuously update the Vendi Score computation, our approach operates with significantly lower memory overhead. (2) Our method eliminates the need to balance between a diversity term and a contextualization term, removing one hyperparameter from the optimization process. This simplification makes our approach more accessible and easier to tune, reducing the complexity of the diversity enhancement pipeline.\looseness-1 %

\input{tables/dp_in1k_indomain}

\subsection{Training downstream image classifiers on synthetic data}

We complement our analysis by evaluating the downstream utility of our Chamfer guidance in the ``static'' ImageNet-1k setup introduced in~\cite{askari25dp}, and generate a dataset of $1,300,000$ synthetic images using our approach, with different $k$ real exemplar images from the training set. Each synthetic image is generated with a simple prompt \texttt{(class name)}. 
We report the accuracy of a ViT-B~\citep{dosovitskiy2021an} classifier trained on this synthetic data and tested on real validation data. \\
We compare the performance of \ldmonefive and \ldmthreefive against a classifier trained on limited real images ($k$ times the number of classes), with results shown in \Cref{tab:choice_of_gens_in1k}. 
We initially train the classifier on only limited real images and present the results in \Cref{tab:choice_of_gens_in1k_real} to serve as reference. Our findings demonstrate that this limited real data alone fails to achieve satisfactory performance. Next, we leverage the available real data per class to generate synthetic images using our Chamfer distance approach, presenting these results in \Cref{tab:choice_of_gens_in1k_mixed}. We study the effect of images generated without and with our Chamfer guidance, and we show that our synthetic-only approach using Chamfer guidance achieves substantial improvements over the base LDM generations, with gains of $+12.01$ and $+15.83$ accuracy, respectively. Finally, we find that combining real and synthetic data yields optimal results, achieving up to $63.81\%$ and $62.61\%$ accuracy when using only 32 real images per class alongside our generated synthetic data using \ldmonefive and \ldmthreefive, respectively. These results also show how our Chamfer guidance can effectively close the gap in grounded diversity between the two models. The difference is more than $5\%$ when using only the synthetic data, and we reduce it to $1.25\%$ with $k=32$. An analogous behavior emerges when leveraging real data, reducing the gap from $3.42$ to $1.20$ with $32$k real images. \\
We also test on ImageNet-V2~\cite{recht2019imagenet}, ImageNet-Sketch~\cite{wang2019learning}, ImageNet-R~\cite{hendrycks2021many}, and ImageNet-A~\cite{hendrycks2021nae} to measure out-of-distribution (OOD) generalization. Models trained with data generated by our Chamfer guidance exhibit strong performance on these out-of-distribution datasets, surpassing models trained on only real data on all variants, given the same number of real images. For ImageNet-Sketch, \ldmthreefive with our Chamfer guidance surpasses the performance of a classifier trained on the full (1.3M samples) ImageNet-1k. Our Chamfer guidance always obtains substantial gains over the default sampling of \ldmonefive and \ldmthreefive across all ImageNet variants, up to $16\%$ when using synthetic data only for ImageNet-Sketch on \ldmthreefive.
Interestingly, on ImageNet-R and ImageNet-A, adding real data to our synthetic ones harms the model's performance. In these domains, which differ significantly from ImageNet-1k, the gain resulting from synthetic data is less pronounced \wrt ImageNet-V2 and ImageNet-Sketch. R(endition) and A(dversarial) represent more challenging domains where more natural-looking images bring reduced benefit. Nevertheless, our Chamfer guidance still surpasses their real-only counterparts. These results confirms the downstream utility of our generated samples for OOD tasks as well. Similarly to the in-domain results, we observe how our Chamfer guidance can effectively close the performance gap between \ldmonefive and \ldmthreefive, and for ImageNet-Sketch, the performance of \ldmthreefive is superior to \ldmonefive.\\
Finally, we would like to observe how the improvement in quality between \ldmonefive and \ldmthreefive does not correspond to a greater utility of the generated data. What we generally observe is a decrease in performance between \ldmonefive and \ldmthreefive, and that our Chamfer guidance can restore the utility of the most recent models.

\subsection{Ablations}

We conduct our ablation studies on \ldmonefive, using the ImageNet-1k dataset. The goal is to understand the robustness of our Chamfer Guidance to the relevant hyperparameters, \ie $\omega$ from \Cref{eq:cfg} (in \Cref{tab:cfg_ablation_sd15_in1k}) and the strength of the Chamfer Guidance $\gamma$ from \Cref{eq:chamfer_guidance} (in the Appendix).\looseness-1 \\
\pp{CFG ablation.} \Cref{tab:cfg_ablation_sd15_in1k} presents the impact of varying the CFG scale $\omega$ on our method.
The results clearly demonstrate that a moderate $\omega$ value of 2.0 with $k=32$ guiding images achieves optimal performance, yielding the highest $F_1$ score (0.931) and competitive precision (0.950) while achieving high coverage (0.912). This balance is crucial for generating both accurate and diverse images. Notably, the FID score of 8.935 at $\omega=2.0$  indicates better image quality compared to higher guidance settings, which deteriorates to $14.388$. This confirms that excessive guidance strength introduces oversaturation artifacts, compromising perceptual quality. The low FDD score (113.301) at $\omega=2.0$ further validates the effectiveness of this configuration in producing faithful, high-quality, and diverse image generations. Interestingly, using the conditional model only ($\omega=1.0$) produces competitive results, confirming that our Chamfer Guidance approach enables state-of-the-art results while reducing computational complexity. We present a more detailed study on efficiency in the Appendix.

\input{tables/cfg_ablation_sd15_in1k}

%% file: tables/in1k_unified_omega.tex
\begin{table}
\caption{Quantitative results on ImageNet-1k using \ldmonefive and \ldmthreefive. 
Our Chamfer guidance consistently achieves state-of-the-art fidelity, diversity, and image quality compared to reference-free and training-based approaches across both models.\looseness-1}
\vspace{-0.75em}
\label{tab:quantitative_combined}
\centering
\begin{subtable}{0.85\textwidth}
\centering
\caption{Results using \ldmonefive.}
\label{tab:quantitative_sd15_in1k}
\footnotesize
\setlength{\tabcolsep}{3.5pt} %
\resizebox{\textwidth}{!}{%
\begin{tabular}{l c c S[table-format=1.3, detect-weight] S[table-format=1.3, detect-weight] S[table-format=1.3, detect-weight] S[table-format=1.3, detect-weight] S[table-format=1.3, detect-weight] S[table-format=3.3, detect-weight] S[table-format=0.3, detect-weight]}
\textbf{Method} & \textbf{$k$} & \textbf{$\omega$} & {\textbf{$\mathbf{F_1}$ (P, C)} $\uparrow$} & {\textbf{Precision} $\uparrow$} & {\textbf{Coverage} $\uparrow$} & {\textbf{Density} $\uparrow$} & {\textbf{Recall} $\uparrow$} & {\textbf{FDD} $\downarrow$} & {\textbf{FID} $\downarrow$} \\
\midrule
\ldmonefive & {--} & 1.0 & 0.507 & 0.723 & 0.391 & 0.551 & 0.656 & 431.241 & 31.289 \\
\rowcolor{gray!10}\ldmonefive & {--} & 7.5 & 0.709 & 0.862 & 0.603 & 0.775 & 0.415 & 248.731 & 16.116 \\
APG~\cite{sadat2024eliminatingoversaturationartifactshigh} & {--} & 4.0 & 0.723 & 0.855 & 0.626 & 0.752 & 0.533 & 217.937 & 13.391 \\
\rowcolor{gray!10}CADS~\cite{sadat2023cads} & {--} & 4.0 & 0.718 & 0.850 & 0.621 & 0.743 & 0.546 & 217.959 & 13.434 \\

Limited Interval~\citep{kynkaanniemi2024applying} & {--} & 4.0 & 0.708 &	0.837 &	0.613 &	0.686 &	0.631 &	219.168 &	11.405 \\
\rowcolor{gray!10}Particle Guidance~\citep{corso2024particle} & {--} & 4.0 & 0.719 &	0.846 &	0.625 &	0.744 &	0.544 &	222.264 &	14.516 \\

\midrule
\rowcolor{gray!10}c-VSG~\cite{hemmat2024improving} & 2 & 2.0 & 0.660 & 0.788 & 0.568 & 0.632 & \textbf{0.738} & 236.337 & 10.742 \\
Vanilla fine-tuning & 2 & 7.5 & 0.733 & 0.867 & 0.635 & 0.797 & 0.495 & 220.089 & 14.869 \\
\rowcolor{gray!10}Vanilla fine-tuning & 32 & 7.5 & 0.741 & 0.869 & 0.646 & 0.796 & 0.503 & 214.045 & 14.723 \\
Chamfer fine-tuning & 2 & 2.0 & 0.759 & 0.886 & 0.664 & 0.868 & 0.420 & 211.714 & 15.464 \\
\rowcolor{gray!10}Chamfer fine-tuning & 32 & 2.0 & 0.766 & 0.898 & 0.668 & 0.875 & 0.404 & 209.999 & 15.492 \\
Chamfer Guidance (Ours) & 2 & 2.0 & 0.886 & 0.947 & 0.833 & 1.108 & 0.480 & 156.179 & 13.670 \\
\rowcolor{gray!10}Chamfer Guidance (Ours) & 32 & 2.0 & {\textbf{0.931}} & {\bfseries 0.950} & \textbf{0.912} & \textbf{1.213} & 0.649 & \textbf{113.301} & \textbf{8.935} \\
\end{tabular}
}
\end{subtable}

\centering
\begin{subtable}{0.85\textwidth}
\centering
\caption{Results using \ldmthreefive.}
\label{tab:quantitative_sd35_in1k}
\footnotesize
\setlength{\tabcolsep}{3.5pt}  %
\resizebox{\textwidth}{!}{%
\begin{tabular}{l c c S[table-format=1.3, detect-weight] S[table-format=1.3, detect-weight] S[table-format=1.3, detect-weight] S[table-format=1.3, detect-weight] S[table-format=1.3, detect-weight] S[table-format=3.3, detect-weight] S[table-format=0.3, detect-weight]}
\textbf{Method} & \textbf{$k$} & \textbf{$\omega$} & {\textbf{$\mathbf{F_1}$ (P, C)} $\uparrow$} & {\textbf{Precision} $\uparrow$} & {\textbf{Coverage} $\uparrow$} & {\textbf{Density} $\uparrow$} & {\textbf{Recall} $\uparrow$} & {\textbf{FDD} $\downarrow$} & {\textbf{FID} $\downarrow$} \\ 
\midrule
\ldmthreefive & {--} & 1.0 & 0.599 & 0.752 & 0.498 & 0.560 & {\bfseries 0.667} & 314.732 & 17.268 \\
\rowcolor{gray!10}\ldmthreefive & {--} & 2.0 & 0.727 & 0.872 & 0.623 & 0.797 & 0.502 & 231.890 & 15.673 \\
APG~\cite{sadat2024eliminatingoversaturationartifactshigh} & {--} & 2.0 & 0.723 & 0.856 & 0.625 & 0.783 & 0.485 &  237.623 & 15.433 \\
\rowcolor{gray!10}CADS~\cite{sadat2023cads} & {--} & 2.0 & 0.717 & 0.851 & 0.620 & 0.749 & 0.518 &  238.125 & 15.139 \\

\midrule
Vanilla fine-tuning & 2 & 4.0 & 0.727 & 0.861 & 0.628 & 0.799 & 0.433 & 238.856 & 18.539 \\
\rowcolor{gray!10}Vanilla fine-tuning & 32 & 4.0 & 0.728 & 0.855 & 0.635 & 0.789 & 0.472 & 232.911 & 18.098 \\
Chamfer fine-tuning & 2 & 2.0 & 0.741 & 0.906 & 0.626 & 0.906 & 0.255 & 284.654 & 21.538 \\
\rowcolor{gray!10}Chamfer fine-tuning & 32 & 2.0 & 0.731 & 0.899 & 0.616 & 0.890 & 0.276 & 286.708 & 22.061 \\
Chamfer Guidance (Ours) & 2 & 1.0 & 0.912 & 0.964 & 0.864 & 1.245 & 0.469 & 134.305 & {\bfseries 8.878} \\
\rowcolor{gray!10}Chamfer Guidance (Ours) & 32 & 1.0 & {\bfseries0.950} & {\bfseries0.975} & {\bfseries 0.927} & \textbf{1.366} & 0.550 & {\bfseries 121.403} & 9.606 \\
\end{tabular}
}
\end{subtable}
\vspace{-2em}
\end{table}

%% file: cvsg_tables/cvsg_geode_pc.tex
\begin{table}[!t]
\vspace{-1em}
\caption{Comparison on the GeoDE dataset under the geographical representation benchmark of c-VSG, with model selection on $F_1 (P,C)$. Metrics are computed in Inception space. $^\dagger$ indicates re-implemented results. AF: Africa, WAS: West Asia, AM: Americas, EU: Europe. ``label'' is region label, ``desc'' is text description, and ``img'' is exemplar images. Our Chamfer Guidance achieves state-of-the-art in terms of $F_1$, and coverage scales with an increased amount of available samples.}
\centering
\small
\setlength{\tabcolsep}{3.5pt}  %
\resizebox{0.9\textwidth}{!}{
\begin{tabular}{
  l
  c
  c %
  c
  *{4}{S[table-format=1.3]}
  *{4}{S[table-format=1.3]}
}
\multirow{2.5}{*}{\textbf{Method}} &
\multirow{2.5}{*}{\textbf{Ref. Info}} &
\multirow{2.5}{*}{$k$} & %
\multirow{2.5}{*}{\textbf{Worst-Reg.}} &
\multicolumn{2}{c}{\textbf{$\mathbf{F_1}$ (P, C)} $\uparrow$} &
\multicolumn{2}{c}{\textbf{Precision} $\uparrow$} &
\multicolumn{2}{c}{\textbf{Coverage} $\uparrow$} &
\multicolumn{2}{c}{\textbf{CLIPScore} $\uparrow$} \\
\cmidrule(lr){5-6} \cmidrule(lr){7-8} \cmidrule(lr){9-10} \cmidrule(lr){11-12} %
& & & & {\textbf{Avg.}} & {\textbf{Worst-Reg.}} & {\textbf{Avg.}} & {\textbf{Worst-Reg.}} & {\textbf{Avg.}} & {\textbf{Worst-Reg.}} & {\textbf{Avg.}} & {\textbf{Worst-Reg.}} \\
\midrule

\recomputed{\ldmonefive$^\dagger$}                                 & \xmark     & -- & AF  & 0.412 & 0.346 & 0.459 & 0.378 & 0.374 & 0.319 & 0.251    & 0.239   \\
\rowcolor{gray!10}Synonyms$^\dagger$                                          & \xmark     & -- & AF  & 0.339 & 0.297 & 0.350 & 0.298 & 0.328 & 0.297 & 0.215 & 0.203 \\

VSG$^\dagger$~\cite{hemmat2024improving}                                    & \xmark     & -- & AF  & 0.353 & 0.312 & 0.349 & 0.307 & 0.357 & 0.317 & 0.180 & 0.191 \\
\midrule
\rowcolor{gray!10}Paraphrasing$^\dagger$                                      & desc       & -- & WAS & 0.329 & 0.301 & 0.338 & 0.309 & 0.320 & 0.293 & 0.231 & 0.228 \\

\recomputed{Semantic Guidance$^\dagger$}                          & label      & -- & AF  & 0.412 & 0.344 & 0.458 & 0.376 & 0.375 & 0.317 & 0.251    & 0.239   \\
\rowcolor{gray!10}\recomputed{FG CLIP (Loss)$^\dagger$}             & label      & -- & AF  & 0.418 & 0.391 & 0.441 & 0.422 & 0.397 & 0.365 & 0.246    & 0.244   \\

\recomputed{FG CLIP (Entropy)$^\dagger$}                             & label      & -- & AF  & 0.414 & 0.357 & 0.437 & 0.393 & 0.393 & 0.327 & 0.238    & 0.236   \\
\rowcolor{gray!10}Textual Inversion~\cite{gal2022imageworthwordpersonalizing}$^\dagger$                     & img        & 4 & EU  & 0.300   &  0.267    & 0.356 & 0.308 & 0.260   & 0.236    & 0.212 & 0.214 \\

c-VSG$^\dagger$ \cite{hemmat2024improving} (CLIP)       & img        & 2 & AM  & 0.435 & 0.412 & 0.424 & 0.408 & 0.446 & 0.416 & 0.254    & 0.254   \\
\rowcolor{gray!10}c-VSG$^\dagger$ \cite{hemmat2024improving} (CLIP)       & img        & 4 & AF  &0.412    & 0.357   & 0.428   &0.382    & 0.398   & 0.335   & 0.253    & 0.253   \\
\midrule

Chamfer Guidance (Ours, CLIP) & img & 2 & AF & 0.495 & 0.461 & 0.525 & 0.487 & 0.468 & 0.437 & {0.265}    & {0.263}   \\
\rowcolor{gray!10}Chamfer Guidance (Ours, CLIP) & img & 4 & AM & 0.492 & 0.457 & 0.513 & 0.449 & \textbf{0.473} & \textbf{0.465} & \textbf{0.267}    & \textbf{0.266}   \\ \addlinespace[3pt]

Chamfer Guidance (Ours, DINOv2) & img & 2 & AF & 0.488 & 0.456 & 0.539 & \textbf{0.513} & 0.446 & 0.411 & {0.257}    & {0.248}   \\
\rowcolor{gray!10}Chamfer Guidance (Ours, DINOv2) & img & 4 & AF & \textbf{0.500} & \textbf{0.469} & \textbf{0.549} & 0.512 & 0.459 & 0.432 & {0.257}    & {0.249}   \\
\end{tabular}
}
 \vspace{-2em}
\label{tab:cvsg_geode_pc}
\end{table}

%% file: tables/dp_in1k_indomain.tex
\begin{table}[htbp]
 \centering
 \vspace{-1em}
 \caption{Validation accuracy on real images of classifiers trained with real data, synthetic data, and a mix of synthetic and training data for ImagetNet-1k. All models are trained for 200k iterations.}
 \label{tab:choice_of_gens_in1k}

 \vspace{-0.75em}
 \resizebox{0.45\linewidth}{!}{
 \subcaptionbox{Real data only.\label{tab:choice_of_gens_in1k_real}}{%
 {\small %
 \begin{tabular}{c S[table-format=2.2] S[table-format=2.2] S[table-format=2.2] S[table-format=2.2] S[table-format=2.2]}
  \textbf{Real images} & {\textbf{IN1k}} & {\textbf{IN-v2}} & {\textbf{IN-Sk}} & {\textbf{IN-R}} & {\textbf{IN-A}} \\
  \midrule
  2k & 5.01 & 3.94 & 0.63 & 0.89 & 0.21 \\
  \rowcolor{gray!10}32k & 34.05 & 25.38 & 4.17 & 5.04 & 0.53 \\
   1.3M & 82.60 & 70.90 & 32.50 & 44.60 & 29.40 \\
 \end{tabular}
 }
 }%
 }
 
 \vspace{0.1em} %
 
 \resizebox{0.9\linewidth}{!}{
 \subcaptionbox{Synthetic only and mixed data. Real and synthetic images refer to the number of samples used to train the classifier.\label{tab:choice_of_gens_in1k_mixed}}{%
 {\small %
 \begin{tabular}{c c c S[table-format=2.2] S[table-format=2.2] S[table-format=2.2] S[table-format=2.2] S[table-format=2.3] S[table-format=2.2] S[table-format=2.2] S[table-format=2.2] S[table-format=2.2] S[table-format=2.3]} %
  \multirow{2}{1cm}{\textbf{Real images}} & \multirow{2}{1cm}{\textbf{Syn. images}} & \multirow{2}{*}{\textbf{Guidance}} & \multicolumn{5}{c}{\textbf{\ldmonefive}} & \multicolumn{5}{c}{\textbf{\ldmthreefive}} \\ %
  \cmidrule(lr){4-8} \cmidrule(lr){9-13} %
  & & & {\textbf{IN1k}} & {\textbf{IN-v2}} & {\textbf{IN-Sk}} & {\textbf{IN-R}} & {\textbf{IN-A}} & {\textbf{IN1k}} & {\textbf{IN-v2}} & {\textbf{IN-Sk}} & {\textbf{IN-R}} & {\textbf{IN-A}} \\ %
  \midrule
  \multirow{3}{*}{0} & \multirow{3}{*}{1.3M} & $\omega=2$ & 47.67 & 40.33 & 20.49 & 17.49 & 1.45 & 37.83 & 34.07 & 17.60 & 11.53 & 0.88 \\
  & & \cellcolor{gray!10}Chamfer k=2 & \cellcolor{gray!10}52.88 & \cellcolor{gray!10}45.37 & \cellcolor{gray!10}28.07 & \cellcolor{gray!10}19.60 & \cellcolor{gray!10}1.71 & \cellcolor{gray!10}52.14 & \cellcolor{gray!10}44.27 & \cellcolor{gray!10}33.47 & \cellcolor{gray!10}20.26 & \cellcolor{gray!10}1.93 \\
  & & Chamfer k=32 & \textbf{54.91} & \textbf{46.43} & \textbf{28.08} & \textbf{19.78} & \textbf{5.11} & \textbf{53.66} & \textbf{45.46} & \textbf{34.44} & \textbf{20.67} & \textbf{5.28} \\
  \cmidrule(lr){1-13} %
  \multirow{2}{*}{2k} & \multirow{2}{*}{1.3M} & $\omega=2$ & 48.47 & 41.07 & 21.21 & 16.96 & 1.57 & 40.89 & 31.72 & 20.03 & 12.31 & 1.32 \\
  & & \cellcolor{gray!10}Chamfer k=2 & \cellcolor{gray!10}\textbf{53.57} & \cellcolor{gray!10}\textbf{46.42} & \cellcolor{gray!10}\textbf{29.48} & \cellcolor{gray!10}\textbf{21.25} & \cellcolor{gray!10}\textbf{1.65} & \cellcolor{gray!10}\textbf{52.95} & \cellcolor{gray!10}\textbf{45.26} & \cellcolor{gray!10}\textbf{33.52} & \cellcolor{gray!10}\textbf{20.51} & \cellcolor{gray!10}\textbf{1.99} \\
  \cmidrule(lr){1-13} %
  \multirow{2}{*}{32k} & \multirow{2}{*}{1.3M} & $\omega=2$ & 59.07 & 49.77 & 25.04 & 20.10 & 2.44 & 55.65 & 45.65 & 21.64 & 14.97 & 1.54 \\
  & & \cellcolor{gray!10}Chamfer k=32 & \cellcolor{gray!10}\textbf{63.81} & \cellcolor{gray!10}\textbf{53.84} & \cellcolor{gray!10}\textbf{32.34} & \cellcolor{gray!10}\textbf{22.40} & \cellcolor{gray!10}\textbf{2.72} & \cellcolor{gray!10}\textbf{62.61} & \cellcolor{gray!10}\textbf{52.58} & \cellcolor{gray!10}\textbf{34.49} & \cellcolor{gray!10}\textbf{21.85} & \cellcolor{gray!10}\textbf{2.36} \\
 \end{tabular}
 }%
 }
 }
 \vspace{-1.5em}
\end{table}

%% file: tables/cfg_ablation_sd15_in1k.tex
\begin{table}[!ht]
    \vspace{-1.em}
    \caption{$\omega$ ablation on ImageNet-1k using \ldmonefive. Our Chamfer Guidance can achieve near state-of-the-art results employing only the conditional model, reducing the needed inference computation.}
    \centering
    \small
    \setlength{\tabcolsep}{4pt}  %
    \resizebox{0.7\textwidth}{!}{
    \begin{tabular}{c c S[table-format=1.3, detect-weight] S[table-format=1.3, detect-weight] S[table-format=1.3, detect-weight] S[table-format=1.3, detect-weight] S[table-format=1.3, detect-weight] S[table-format=3.3, detect-weight] S[table-format=2.3, detect-weight]}
    $\mathbf{\omega}$ & \textbf{$k$} & {\textbf{$\mathbf{F_1}$ (P, C)} $\uparrow$} & {\textbf{Precision} $\uparrow$} & {\textbf{Coverage} $\uparrow$} & {\textbf{Density} $\uparrow$} & {\textbf{Recall} $\uparrow$} & {\textbf{FDD} $\downarrow$} & {\textbf{FID} $\downarrow$} \\\midrule
    1.0 & 2 & 0.849 & 0.890 & 0.811 & 0.904 & \textbf{0.736} & 150.748 & 13.217 \\
    \rowcolor{gray!10}1.0 & 32 & 0.899 & 0.923 & 0.876 & 1.086 & 0.735 & 117.834 & 9.759 \\
    2.0 & 2 & 0.881 & 0.932 & 0.835 & 1.051 & 0.637 & 124.191 & 8.840 \\
    \rowcolor{gray!10}2.0 & 32 & \textbf{0.931} & 0.950 & \textbf{0.912} & 1.213 & 0.649 & \textbf{113.301} & \textbf{8.935} \\
    7.5 & 2 & 0.886 & 0.947 & 0.833 & 1.108 & 0.480 & 156.179 & 13.670 \\
    \rowcolor{gray!10}7.5 & 32 & 0.925 & \textbf{0.957} & 0.894 & \textbf{1.238} & 0.498 & 153.111 & 14.388 \\
    \end{tabular}}
    \vspace{-2em}
    \label{tab:cfg_ablation_sd15_in1k}
    \end{table}

%% file: sec/5_conclusions.tex
\section{Conclusion}
\label{sec:conclusions}

{
\setlength{\parskip}{0pt}

We introduced Chamfer Guidance, a novel training-free approach to improve the utility of synthetic data from conditional image models. By leveraging a small set of exemplar real images to guide the generation process, our method balances quality and diversity while addressing distribution shifts between synthetic and real data. Experimental results on ImageNet-1k and geo-diversity benchmarks show Chamfer Guidance achieves state-of-the-art performance and scales its effectiveness as exemplars increase. Additionally, our synthetic data can be used in downstream applications. Furthermore, our approach eliminates the computational overhead of CFG, reducing computational requirements while maintaining superior performance in quality and diversity. These contributions advance synthetic data generation and demonstrate that thoughtfully designed guidance can unlock the full potential of conditional image generative models for downstream applications.

\pp{Limitations.} Our evaluation of reference-based diversity and quality is built on automated metrics, which inherit intrinsic biases, \eg, the distribution of the dataset we compare against, or the reliance on pre-existing feature extractors that might not capture subtle differences. While we rely on previous works to provide the most accurate metrics possible, these are statistical aggregators and do not account for individual preferences. Currently, our approach is designed for class-conditional models and does not directly support text-to-image generation.
This limitation opens up several exciting avenues for future work. To extend our method to text-to-image models, we envision a retrieval-based pipeline. First, a large text-image dataset would be embedded into semantic vectors to create an offline retrieval database. Then, at inference time, for a user's text query, the top-k relevant images would be retrieved to serve as exemplars for our guidance. Another interesting extension would be a zero-shot, data-free pipeline. We could envision a self-bootstrapping technique that first generates initial candidate images for a class, then automatically selects a diverse subset maximizing coverage (or ``diameter'') in a robust feature space, and finally uses this synthetically-generated set as the guidance exemplars.\looseness-1

\pp{Societal Impact.} Our work builds on image generative models, and it inherits some of the societal challenges of image content creation. However, our Chamfer Guidance provides a new way to guide the generations towards a set of exemplary images. As such, it equips the user with a broader inference-time tool set to better control the sampling process.

\pp{Acknowledgments.} This work was sponsored by the project FAIR Future AI Research
(PE00000013), funded by NextGeneration EU.
}

%% file: sec/10_appendix.tex
\clearpage

\appendix

\input{sec/appendices/efficiency}

\input{sec/appendices/evaluation_metrics}

\input{sec/appendices/user_study}

\input{sec/appendices/cvsg_baselines}

\input{sec/appendices/cvsg_additional_results}

\input{sec/appendices/ablations}

\input{sec/appendices/obj_additional_results}

\input{sec/appendices/dp_additional_results}

\input{sec/appendices/licenses}

\input{sec/appendices/qualitatives}

%% file: sec/appendices/efficiency.tex
\section{Efficiency analysis}

We analyze the computational overhead and efficiency gains of our proposed Chamfer guidance method compared to traditional Classifier-Free Guidance (CFG) when applied to \ldmonefive and \ldmthreefive. This analysis quantifies the significant computational savings achieved by our approach. We use \texttt{pytorch's FlopCounterMode} to compute the FLOPs of each module. 

\subsection{Baseline Diffusion Model Computation}

\pp{\ldmonefive.} The standard diffusion process using \ldmonefive requires $\approx 800 \text{GFLOPs}$ per forward pass. With the default implementation using 40 denoising steps, the total computation for generating a single sample amounts to $40 \times 800 \text{ GFLOPs} = 32 \text{ TFLOPs}$.

When applying CFG, which requires a doubled batch size per step, the computational requirements double, $40 \times 1.6 \text{ TFLOPs} = 64 \text{ TFLOPs}$.

\pp{\ldmthreefive.} The standard diffusion process using \ldmthreefive requires approximately 6 TFLOPs per forward pass. With the implementation using 40 denoising steps, the total computation for generating a single sample amounts to $40 \times 6 \text{ TFLOPs} = 240 \text{ TFLOPs}$.

When applying CFG, which requires a doubled batch size per step, the computational requirements double, $40 \times 12 \text{ TFLOPs} = 480 \text{ TFLOPs}$.

We also have to take into consideration the decoding of the final samples, which account for $\approx 2 \text{TFlops}$ for \ldmonefive, and $\approx 10 \text{TFlops}$ for \ldmthreefive.

Therefore we estimate $34$ and $66$ TFLOPs to generate an image without and with CFG for \ldmonefive, and $250$ and $490$ TFLOPs to generate an image without and with CFG for \ldmthreefive. 

\subsection{Our Chamfer Guidance Approach}

The Chamfer guidance method introduces two additional computational components:

\paragraph{Reference Image Processing} We use DINOv2 (ViT/L) for feature extraction, which requires $\approx~160$ GFLOPs per image. Given our largest reference set of $k=32$ real images, the total computation for reference embedding is: $32 \times 160 \text{ GFLOPs} \approx 5.1 \text{ TFLOPs}$.
This is a one-time cost incurred at the beginning of the sampling process.

During generation, we apply our guidance at a frequency of $G_{\text{freq}} = 5$, meaning we compute the guidance every 5 steps. Each guidance computation using DINOv2 requires first decoding the samples, and then encoding in DINO space, therefore the cost for \ldmonefive is $(2 \text{TFLOPs} + 160 \text{ GFLOPs}) \times 8 \approx 17.3 \text{ TFLOPs}$, while for \ldmthreefive is: $(10 \text{TFLOPs} + 160 \text{ GFLOPs}) \times 8 \approx 81 \text{ TFLOPs}$.

For the entire sampling process with 40 steps, the reference encoding and the Chamfer guidance adds approximately $\approx 22.4 \text{ TFLOPs}$ for \ldmonefive, and  $\approx 86 \text{ TFLOPs}$ for \ldmthreefive.

The total computation required by our Chamfer guidance method is therefore about $56.4 \text{ TFLOPs}$ for \ldmonefive, and $336 \text{ TFLOPs}$ for \ldmthreefive.

\subsection{Efficiency Comparison}

Comparing our method with the standard CFG approach:

\begin{align}
\text{Efficiency gain \ldmonefive} &= 1 - \frac{\text{Computation}_{\text{Chamfer}}}{\text{Computation}_{\text{CFG}}} \\
&= 1 - \frac{56.4 \text{ TFLOPs}}{66 \text{ TFLOPs}} \\
&\approx 0.15 \text{ or } 15\%
\end{align}

\begin{align}
\text{Efficiency gain \ldmthreefive} &= 1 - \frac{\text{Computation}_{\text{Chamfer}}}{\text{Computation}_{\text{CFG}}} \\
&= 1 - \frac{336 \text{ TFLOPs}}{490 \text{ TFLOPs}} \\
&\approx 0.31 \text{ or } 31\%
\end{align}

This analysis demonstrates that our Chamfer guidance method achieves superior generation quality while reducing the computational requirements by approximately 31\% compared to traditional CFG on \ldmthreefive. This translates to $\approx 4$s. to generate a sample with \ldmthreefive on a RTX A6000.

%% file: sec/appendices/evaluation_metrics.tex
\section{On the evaluation metrics}

Through our experimentation, we identified several limitations in commonly used metrics for evaluating generative diversity, leading us to choose the make the following choices to the evaluation framework:

\pp{From Recall to Coverage:} We examine the use of recall as a grounded diversity metric and find it to be highly sensitive to outliers in the generated distribution~\cite{naeem2020reliable}. A single generated sample that happens to be far from the other generated points can disproportionately inflate the recall value. As an alternative, we advocate for the use of coverage, which provides a more reliable assessment of how well the generated distribution matches the target diverse distribution. Coverage is more robust to outliers, as the manifold is computed \wrt real points~\cite{naeem2020reliable}.

\pp{$\mathbf{F_1}$ with precision and coverage:} To evaluate both fidelity and diversity in a single metric, we propose to modify the $F_1$ score in previous works~\cite{hemmat2024improving} to harmonically combine precision and coverage. We specifically chose precision over density~\cite{naeem2020reliable} despite their similar purposes, as precision is naturally bounded between 0 and 1, making it more intuitive to interpret and combine with other metrics. In contrast, density is potentially unbounded, which complicates its use in composite metrics.

\pp{From Inception to DINO space for evaluation:} We assess the choice of feature space for computing similarity-based metrics. While the Inception feature space has been the \textit{de facto} standard in generative model evaluation, we advocate for the use of the DINO feature space instead, as supported by previous works~\cite{stein2023exposing,hall2024towards}. Self-supervised latent spaces better align with human perception of image realism, while Inception focuses on specific objects rather than holistic image features, often ignoring important aspects of images.

%% file: sec/appendices/user_study.tex
\section{User Study}

While the primary focus of our study is the utility of the generated data for representation and as a training source for downstream tasks, which is confirmed by our extensive experimental evaluation with downstream classifier training, we also evaluate human perception of our generated images.

\begin{figure*}[!th]
    \centering
    \includegraphics[width=0.75\linewidth]{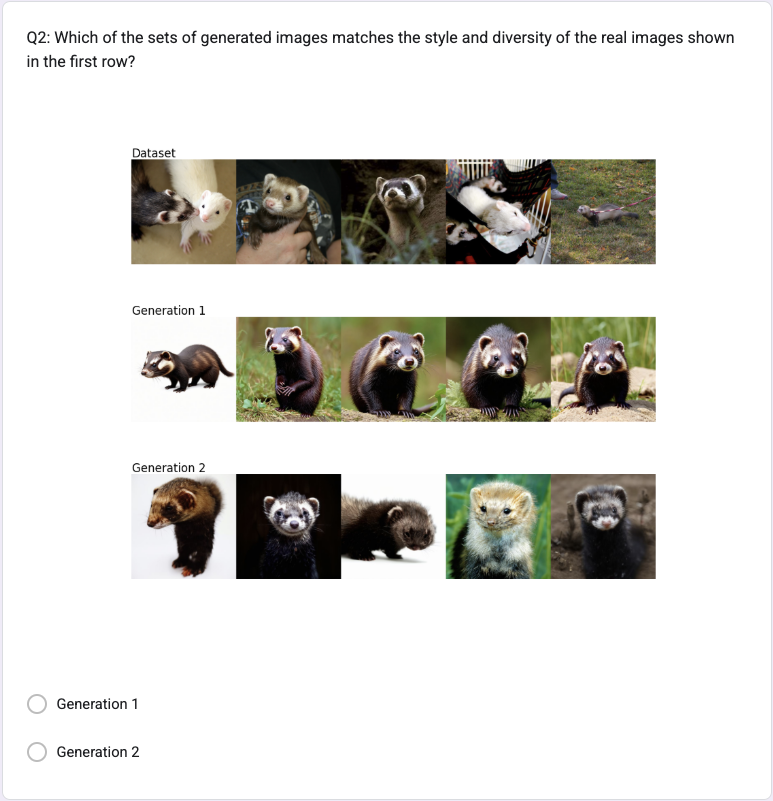}
    \caption{Example of user study question.}
    \label{fig:user_study}
\end{figure*}

To this end, we conduct a small-scale user study to complement our quantitative findings. We collected 965 data points from more than twenty anonymous annotators. In this study, users were presented with samples generated from prompts based on the ImageNet dataset. Their task was to choose their preferred generation in a side-by-side comparison between images from the base \ldmthreefive model and images generated with our Chamfer Guidance applied to the same model. Users were also presented with real images from the dataset to ground their evaluation in real-world quality and coherence, as we show in \Cref{fig:user_study}.

The results showed a strong preference for our method: images generated with our Chamfer Guidance were preferred in $\mathbf{92\% \pm 2\%}$ of the cases. This suggests that the automatic evaluation of quantitative improvements in downstream utility also correlates with enhanced human-perceived quality and fidelity to the target concept distribution.

%% file: sec/appendices/cvsg_baselines.tex
\section{Geographic diversity baselines}

We employ the same baselines defined in c-VSG~\cite{hemmat2024improving}, which uses reference-free and reference-based solutions. We report descriptions for them as follows.

\noindent\textbf{Without any additional information:} 
\begin{itemize}[leftmargin=*]
    \item \ldmonefive: This is the baseline setup where the base \ldmonefive is used with the prompt 
    \objreg to condition the generation process.
    \item \texttt{Synonyms}: This strategy maps each object class to its corresponding ImageNet~\cite{deng2009imagenet} class and WordNet~\cite{miller1995wordnet} synset. 
    For each class, we generate images that cover all possible meanings (lemmas), including the original object word. This means that each synset, which groups lemmas for a specific sense of the class, guides the image generation process through \synreg.
\end{itemize}
\noindent\textbf{With additional information:} 
\begin{itemize}[leftmargin=*]
    \item \texttt{Paraphrasing}: LLaMA-2-70B-chat \cite{touvron2023llama} large language model is used to generate paraphrases of the original prompt template, \objreg. 
    The authors include the specifications and descriptions used in the collection of GeoDE and DollarStreet.
    The metaprompts and paraphrases, as well as the method of tuning prompts and model specifications, are included in the original paper Supplementary material~\cite{hemmat2024improving}. 
    \item \texttt{Semantic Guidance}: 
    Generated images often exhibit diversity problems, as highlighted by previous works~\cite{hall2023dig}, due to an amplification of region-specific object features that go beyond what is present in the evaluation task. To mitigate this overemphasis on regional data, the authors utilized Semantic Guidance~\cite{brack2023sega,friedrich2023fair}, which involves applying negative guidance corresponding to the regional term for each generated image.
    
    \item \texttt{Feedback Guidance (FG)}: 
    Inspired by ~\cite{hemmat2023feedbackguided,sehwag2022generating}, the authors employ an external CLIP-based classifier ~\cite{radford2021learning} to provide feedback during image generation by predicting region labels of these images. To promote greater diversity in the generated outputs, they investigate two feedback guidance strategies: the first maximizes the \texttt{loss} of the classifier, while the second maximizes the \texttt{entropy} of its predicted class distributions.
    
    \item \texttt{Textual Inversion}: 
    Textual Inversion is a technique used to teach text-to-image models new, specific visual concepts from just a few example images. It achieves this by creating a new ``pseudo-word'' or token in the model's vocabulary that becomes associated with the visual characteristics of the provided images.
    We apply textual inversion by learning an embedding for each object in the dataset using four images per object.
\end{itemize}

%% file: sec/appendices/cvsg_additional_results.tex
\section{Additional geographic diversity results}

In \Cref{tab:cvsg_geode_pc_dino} we report the same experiments on GeoDE as in \Cref{tab:cvsg_geode_pc} of the main body, but with metrics in DINOv2~\cite{oquab2023dinov2} space, which is known to align better with human evaluation~\cite{stein2023exposing,hall2024towards}. We observe state-of-the-art results \wrt relevant baselines, and in particular in terms of $F_1$ by a significant margin ($+12\%$) over c-VSG~\cite{hemmat2024improving}, due to an increased precision ($+26.7\%$) and coverage ($+7.7\%$). The results reported here present a different scale \wrt the object-centric scenario due to the $k$ of the kNN computation of metrics, which we maintain equal to c-VSG benchmark, and set it to 3, differently from the default value of 5 when using \texttt{dgm-eval}~\cite{stein2023exposing}.

\input{cvsg_tables/cvsg_geode_pc_dino}

In \Cref{tab:cvsg_dollarstreet_pc} we present the results computed in InceptionV3 space for DollarStreet~\cite{gaviria2022dollar}, using the $F_1$ selection on precision and coverage. Our Chamfer guidance outperforms c-VSG~\cite{hemmat2024improving} by $3.5\%$ in terms of $F_1$ due to both an increased fidelity (precision) and diversity (coverage) of the generated samples. As previously reported for GeoDE, our method exhibits an increased image-text alignment when using CLIP as the feature extractor for the guidance.

\input{cvsg_tables/cvsg_dollarstreet_pc}

In \Cref{tab:cvsg_geode_pr} and \Cref{tab:cvsg_dollarstreet_pr} we report the result with selection of the $F_1$ score between precision and recall, as in the original c-VSG benchmark. Although recall does not represent the best metric for evaluating grounded diversity~\cite{naeem2020reliable}, our Chamfer guidance surpasses c-VSG by $3.9\%$ in terms of recall in GeoDE, and by $2.7\%$ on DollarStreet, leading to state-of-the-art $F_1$ in both datasets. Interestingly, when using recall, grounding the generation on more samples does not lead to increased diversity with CLIP, and to a marginal one with DINOv2. This was the case when using coverage, and we consider these results to be a result of the inflated recall manifold, which promotes outlier generated images.

\input{cvsg_tables/cvsg_geode_pr}

\input{cvsg_tables/cvsg_dollarstreet_pr}

%% file: cvsg_tables/cvsg_geode_pc_dino.tex
\begin{table}[!ht]
\caption{Comparison on the GeoDE dataset under the geographical representation benchmark of c-VSG setting, with model selection on $F_1 (P,C)$. Metrics are computed in DINOv2 space. All the results are re-implemented. AF: Africa, AS: Asia, EU: Europe. ``label'' refers to region label, ``desc'' to text description, and ``img'' to exemplar images.}
\centering
\small
\setlength{\tabcolsep}{4pt}  %
\resizebox{1.\textwidth}{!}{

\begin{tabular}{
  l %
  c %
  c %
  c %
  *{4}{S[table-format=1.3]} %
  *{4}{S[table-format=1.3]} %
}
\multirow{2.5}{*}{\textbf{Method}} & 
\multirow{2.5}{*}{\textbf{Ref. Info}} & 
\multirow{2.5}{*}{\textbf{$k$}} & %
\multirow{2.5}{*}{\textbf{Worst-Reg.}} & 
\multicolumn{2}{c}{\textbf{$\mathbf{F_1}$ (P, C) $\uparrow$}} & 
\multicolumn{2}{c}{\textbf{Precision} $\uparrow$} & 
\multicolumn{2}{c}{\textbf{Coverage} $\uparrow$} & 
\multicolumn{2}{c}{\textbf{CLIPScore} $\uparrow$} \\
\cmidrule(lr){5-6} \cmidrule(lr){7-8} \cmidrule(lr){9-10} \cmidrule(lr){11-12} %
& & & & %
{\textbf{Avg.}} & {\textbf{Worst-Reg.}} & 
{\textbf{Avg.}} & {\textbf{Worst-Reg.}} & 
{\textbf{Avg.}} & {\textbf{Worst-Reg.}} & 
{\textbf{Avg.}} & {\textbf{Worst-Reg.}} \\
\midrule
\rowcolor{gray!10}
\recomputed{\ldmonefive}                                & \xmark     & -- & AF & 0.166 & 0.126 & 0.345 & 0.284 & 0.109 & 0.081 & 0.244   & 0.233\\ %
\midrule

\recomputed{Semantic Guidance}        & label & -- & AF & 0.160 & 0.117 & 0.345 & 0.289 & 0.104 & 0.073 &  0.248   & 0.234   \\ %
\rowcolor{gray!10}
\recomputed{FG CLIP (Loss)}           & label & -- & AF & 0.186 & 0.154 & 0.339 & 0.308 & 0.129 & 0.103 & 0.246 & 0.244   \\ %

\recomputed{FG CLIP (Entropy)}        & label & -- & AF & 0.178 & 0.133 & 0.328 & 0.272 & 0.122 & 0.088 & 0.238  & 0.236   \\ %
\rowcolor{gray!10}
\recomputed{c-VSG}~\cite{hemmat2024improving} (CLIP)    & img   & 2 & AF & 0.183 & 0.139 & 0.337 & 0.288 & 0.126 & 0.091 & 0.254  & 0.252   \\ %
\recomputed{c-VSG} \cite{hemmat2024improving} (CLIP)                           & img   & 4 & AF & 0.184   & 0.133   & 0.338   & 0.289   & 0.127   & 0.086   & \textbf{0.254}  & \textbf{0.253}   \\ %
\midrule
Chamfer Guidance (Ours, DINOv2)          & img   & 2 & AF & 0.246 & 0.221 & 0.551 & 0.507 & 0.159 & 0.141 & 0.248   & 0.242   \\ %
\rowcolor{gray!10}
Chamfer Guidance (Ours, DINOv2)          & img   & 4 & AF & \textbf{0.304} & \textbf{0.292} & \textbf{0.605} & \textbf{0.564} & \textbf{0.204} & \textbf{0.197} & 0.250   & 0.244   \\ %
\end{tabular}
}
\label{tab:cvsg_geode_pc_dino}
\end{table}

%% file: cvsg_tables/cvsg_dollarstreet_pc.tex
\begin{table}[!ht]
\caption{Comparison on the DollarStreet dataset under the geographical representation benchmark of c-VSG setting, with model selection on $F_1 (P,C)$. Metrics are computed in Inception space. $^\dagger$ indicates re-implemented results. AF: Africa, AS: Asia, EU: Europe. ``label'' refers to region label, ``desc'' to text description, and ``img'' to exemplar images. Our Chamfer guidance achieves state-of-the-art in terms of $F_1$, obtaining the best grounded diversity.}
\centering
\small
\setlength{\tabcolsep}{4pt}  %
\resizebox{1.\textwidth}{!}{
\begin{tabular}{
  l
  cc
  c
  *{4}{S[table-format=1.3, detect-weight]}
  *{4}{S[table-format=1.3, detect-weight]}
}
\multirow{2.5}{*}{\textbf{Method}} & 
\multirow{2.5}{*}{\textbf{Ref. Info}} & 
\multirow{2.5}{*}{${k}$} & 
\multirow{2.5}{*}{\textbf{Worst-Reg.}} & 
\multicolumn{2}{c}{\textbf{$\mathbf{F_1}$ (P, C)} $\uparrow$} & 
\multicolumn{2}{c}{\textbf{Precision} $\uparrow$} & 
\multicolumn{2}{c}{\textbf{Coverage} $\uparrow$} & 
\multicolumn{2}{c}{\textbf{CLIPScore} $\uparrow$} \\
\cmidrule(lr){5-6} \cmidrule(lr){7-8} \cmidrule(lr){9-10} \cmidrule(lr){11-12}
& & & & {\textbf{Avg.}} & {\textbf{Worst Reg.}} & {\textbf{Avg.}} & {\textbf{Worst Reg.}} & {\textbf{Avg.}} & {\textbf{Worst Reg.}} & {\textbf{Avg.}} & {\textbf{Worst Reg.}} \\
\midrule
\rowcolor{gray!10}
\recomputed{\ldmonefive$^\dagger$}                                         & \xmark & -- & AS & 0.473 & 0.445 & 0.504 & 0.504 & 0.447 & 0.398 & 0.249 & \textbf{0.255}   \\
Synonyms$^\dagger$                                    & \xmark & -- & AS & 0.445   & 0.435 & 0.448 & 0.439 &  0.443   & 0.432   & 0.216 & 0.219 \\
\rowcolor{gray!10}
VSG$^\dagger$ \cite{hemmat2024improving}                               & \xmark & -- & AS & 0.424   & 0.404   & 0.421 & 0.419 & 0.428 &0.390    & 0.195 & 0.194 \\
\midrule
Paraphrasing$^\dagger$                                & desc & -- & AF & 0.440   & 0.436   & 0.445 & 0.451 & 0.436   & 0.422   & 0.226 & 0.215 \\
\rowcolor{gray!10}
\recomputed{Semantic Guidance$^\dagger$}        & label & -- & AS & 0.473 & 0.446 & 0.504 & 0.507 & 0.447 & 0.398 & 0.249   & \textbf{0.255}  \\
\recomputed{FG CLIP (Loss)$^\dagger$}           & label & -- & AS & 0.468 & 0.429 & 0.481 & 0.461 & 0.456 & 0.401 & 0.243   & 0.244   \\
\rowcolor{gray!10}
\recomputed{FG CLIP (Entropy)$^\dagger$}        & label & -- & AS & 0.468 & 0.440 & 0.487 & 0.500 & 0.453 & 0.393 & \textbf{0.245}  & 0.253   \\
Textual Inversion$^\dagger$                          & img & 4 & AS & 0.076   & 0.037   & 0.505 & 0.491 & 0.042   & 0.019   & 0.213 & 0.216 \\
\rowcolor{gray!10}
c-VSG$^\dagger$ \cite{hemmat2024improving} (CLIP)                            & img & 4 & EU & 0.517 & 0.504 & 0.510 & 0.481 & 0.526 & \textbf{0.529} & 0.241   & 0.241   \\
\midrule
Chamfer Guidance (Ours, CLIP)  & img & 4 & AS & 0.528 & 0.512 & 0.515 & 0.514 & 0.542 & 0.510 & {0.244}   & {0.245}   \\
\addlinespace[3pt]
\rowcolor{gray!10}
Chamfer Guidance (Ours, DINOv2) & img & 4 & AS & \textbf{0.552} & \textbf{0.532} & \textbf{0.545} & \textbf{0.542} & \textbf{0.560} & 0.520 & {0.240}   & {0.244}   \\
\end{tabular}
}
\label{tab:cvsg_dollarstreet_pc}
\end{table}

%% file: cvsg_tables/cvsg_geode_pr.tex
\begin{table}[ht]
\caption{
Comparison on the GeoDE dataset under the geographical representation benchmark of c-VSG setting, with model selection on $F_1 (P,R)$. Metrics are computed in Inception space. AF: Africa, AS: Asia, EU: Europe. ``label'' refers to region label, ``desc'' to text description, and ``img'' to exemplar images.
}
\centering
\small
\setlength{\tabcolsep}{4pt}
\resizebox{1.\textwidth}{!}{
\begin{tabular}{
  l
  cc
  c
  *{4}{S[table-format=1.3, detect-weight]}
  *{4}{S[table-format=1.3, detect-weight]}
}
\multirow{2.5}{*}{\textbf{Method}} & 
\multirow{2.5}{*}{\textbf{Ref. Info}} & 
\multirow{2.5}{*}{${k}$} & 
\multirow{2.5}{*}{\textbf{Worst-Reg.}} & 
\multicolumn{2}{c}{\textbf{$\mathbf{F_1}$ (P, R)} $\uparrow$} & 
\multicolumn{2}{c}{\textbf{Precision} $\uparrow$} & 
\multicolumn{2}{c}{\textbf{Recall} $\uparrow$} & 
\multicolumn{2}{c}{\textbf{CLIPScore} $\uparrow$} \\
\cmidrule(lr){5-6} \cmidrule(lr){7-8} \cmidrule(lr){9-10} \cmidrule(lr){11-12}
& & & & {\textbf{Avg.}} & {\textbf{Worst Reg.}} & {\textbf{Avg.}} & {\textbf{Worst Reg.}} & {\textbf{Avg.}} & {\textbf{Worst Reg.}} & {\textbf{Avg.}} & {\textbf{Worst Reg.}} \\
\midrule
\rowcolor{gray!10}
\ldmonefive                                         & \xmark & -- & AF & 0.364 & 0.322 & 0.413 & 0.273 & 0.337 & 0.395 & 0.242 & 0.218 \\
Synonyms                                              & \xmark & -- & AF & 0.357 & 0.306 & 0.350 & 0.298 & 0.366 & 0.315 & 0.215 & 0.203 \\
\rowcolor{gray!10}
VSG \cite{hemmat2024improving}             & \xmark & -- & AF & 0.399 & 0.356 & 0.349 & 0.307 & 0.470 & 0.424 & 0.180 & 0.191 \\
\midrule
Paraphrasing                                          & desc & -- & WAS & 0.384 & 0.354 & 0.338 & 0.309 & 0.449 & 0.415 & 0.231 & 0.228 \\
\rowcolor{gray!10}
Semantic Guidance                          & label & -- & WAS & 0.420 & 0.401 & \textbf{0.459} & \textbf{0.519} & 0.391 & 0.326 & \textbf{0.245} & 0.253 \\
FG CLIP (Loss)                              & label & -- & WAS & 0.409 & 0.378 & 0.387 & 0.383 & 0.436 & 0.373 & 0.228 & 0.223 \\
\rowcolor{gray!10}
FG CLIP (Entropy)                           & label & -- & AF & 0.380 & 0.337 & 0.340 & 0.329 & 0.429 & 0.345 & 0.224 & 0.227 \\
Textual Inversion                                     & img & 4 & AF & 0.369 & 0.363 & 0.409 & 0.444 & 0.338 & 0.308 & 0.234 & 0.232 \\
\rowcolor{gray!10}
c-VSG \cite{hemmat2024improving} (CLIP)          & img & 2 & AF & 0.455 & 0.444 & 0.424 & 0.417 & 0.493 & 0.476 & 0.254 & 0.253 \\
\midrule
Chamfer Guidance (Ours, CLIP)              & img & 2 & AF & 0.454 & 0.440 & 0.398 & 0.401 & \textbf{0.532} & 0.489 & 0.245 & 0.245 \\
\rowcolor{gray!10}
Chamfer Guidance (Ours, CLIP)              & img & 4 & AF & \textbf{0.463} & \textbf{0.449} & 0.427 & 0.413 & 0.509 & \textbf{0.492} & 0.251 & 0.246 \\
Chamfer Guidance (Ours, DINOv2)              & img & 2 & AF & 0.451 & 0.428 & 0.435 & 0.402 & 0.470 & 0.457 & 0.235 & 0.225 \\
\rowcolor{gray!10}
Chamfer Guidance (Ours, DINOv2)              & img & 4 & AF & 0.460 & 0.437 & 0.453 & 0.401 & 0.472 & 0.481 & 0.236 & 0.225 \\
\end{tabular}
}
\label{tab:cvsg_geode_pr}
\end{table}

%% file: cvsg_tables/cvsg_dollarstreet_pr.tex
\begin{table}[!ht]
\caption{Comparison on the DollarStreet dataset under the geographical representation benchmark of c-VSG setting, with model selection on $F_1 (P,R)$. Metrics are computed in Inception space. $^\dagger$ indicates re-implemented results. AF: Africa, AS: Asia, EU: Europe. ``label'' refers to region label, ``desc'' to text description, and ``img'' to exemplar images.}
\centering
\small
\setlength{\tabcolsep}{4pt}
\resizebox{1.\textwidth}{!}{
\begin{tabular}{
  l
  cc
  c
  *{4}{S[table-format=1.3, detect-weight]}
  *{4}{S[table-format=1.3, detect-weight]}
}
\multirow{2.5}{*}{\textbf{Method}} & 
\multirow{2.5}{*}{\textbf{Ref. Info}} & 
\multirow{2.5}{*}{${k}$} & 
\multirow{2.5}{*}{\textbf{Worst-Reg.}} & 
\multicolumn{2}{c}{\textbf{$\mathbf{F_1}$ (P, R)} $\uparrow$} & 
\multicolumn{2}{c}{\textbf{Precision} $\uparrow$} & 
\multicolumn{2}{c}{\textbf{Recall} $\uparrow$} & 
\multicolumn{2}{c}{\textbf{CLIPScore} $\uparrow$} \\
\cmidrule(lr){5-6} \cmidrule(lr){7-8} \cmidrule(lr){9-10} \cmidrule(lr){11-12}
& & & & {\textbf{Avg.}} & {\textbf{Worst Reg.}} & {\textbf{Avg.}} & {\textbf{Worst Reg.}} & {\textbf{Avg.}} & {\textbf{Worst Reg.}} & {\textbf{Avg.}} & {\textbf{Worst Reg.}} \\
\midrule
\rowcolor{gray!10}
\ldmonefive                                         & \xmark & -- & AS & 0.448 & 0.442 & 0.428 & 0.434 & 0.472 & 0.450 & 0.231 & 0.235 \\
Synonyms                                    & \xmark & -- & AS & 0.464 & 0.457 & 0.451 & 0.448 & 0.467 & 0.467 & 0.216 & 0.220 \\
\rowcolor{gray!10}
VSG \cite{hemmat2024improving}             & \xmark & -- & AS & 0.457 & 0.444 & 0.413 & 0.388 & 0.516 & 0.518 & 0.191 & 0.198 \\
\midrule
Paraphrasing                                & desc & -- & AF & 0.454 & 0.445 & 0.445 & 0.454 & 0.465 & 0.437 & 0.226 & 0.215 \\
\rowcolor{gray!10}
Semantic Guidance                           & label & -- & AS & 0.470 & 0.458 & 0.447 & 0.449 & 0.467 & 0.467 & 0.230 & 0.233 \\
FG CLIP (Loss)                              & label & -- & AS & 0.437 & 0.394 & 0.401 & 0.321 & 0.488 & 0.510 & 0.223 & 0.206 \\
\rowcolor{gray!10}
FG CLIP (Entropy)                           & label & -- & AS & 0.465 & 0.462 & 0.412 & 0.404 & 0.535 & \textbf{0.540} & 0.222 & 0.219 \\
Textual Inversion              & img & 4 & AS & 0.425 & 0.398 & 0.478 & 0.491 & 0.386 & 0.335 & 0.217 & 0.219 \\
\rowcolor{gray!10}
c-VSG$^\dagger$ (CLIP)~\cite{hemmat2024improving} & img & 4 & AS & 0.497 & 0.483 & 0.486 & 0.486 & 0.511 & 0.479 & 0.234 & 0.238 \\
\midrule
Chamfer Guidance (Ours, CLIP)    & img & 4 & AF & 0.492 & 0.478 & 0.464 & 0.458 & 0.524 & 0.498 & 0.231 & 0.225 \\
\rowcolor{gray!10}
Chamfer Guidance (Ours, DINOv2)    & img & 4 & AF & \textbf{0.508} & \textbf{0.484} & \textbf{0.482} & \textbf{0.483} & \textbf{0.538} & 0.486 & 0.225 & 0.215 \\
\end{tabular}
}
\label{tab:cvsg_dollarstreet_pr}
\end{table}

%% file: sec/appendices/ablations.tex
\section{Additional ablation}

\paragraph{$\gamma$ ablation.}

This ablation examines the performance using different Chamfer guidance strengths ($\gamma$). The results in \Cref{tab:str_ablation_sd15_in1k} show that this hyperparameter significantly impacts model performance, with $\gamma = 0.07$ and $k = 32$ achieving the best overall results. This configuration yields the highest $F_1$ score (0.931) and precision (0.950), indicating superior fidelity and diversity. Interestingly, while stronger guidance generally improves these metrics, there is a trade-off with image quality, the FID score increases from 8.840 at $\gamma = 0.05$ to 13.670 at $\gamma = 0.07$. This suggests that while stronger Chamfer guidance ($\gamma = 0.07$) produces more accurate samples, it somewhat compromises the distribution statistics compared to the moderate guidance setting ($\gamma = 0.05$), which maintains a better balance between precision/coverage performance and image quality. This might also be due to the use of the Inception network for the computation of FID instead of DINO.\looseness-1

\input{tables/strength_ablation_sd15_in1k}

\paragraph{Feature extractor ablation.}

To further validate the choice of DINOv2, we conducted a preliminary empirical study comparing its performance as a feature extractor for performing Chamfer guidance against CLIP on the GeoDE dataset in an ``object-centric'' setting, \eg, with prompts like \texttt{"a photo of a car."}. These experiments were run using \ldmonefive.

Our findings were as follows:

\begin{itemize}
\item DINOv2 consistently yielded the best performance, showing substantial gains in both diversity and fidelity. Similarity in DINOv2 space also correlates to human-perceived similarity slightly more than CLIP, as reported in \cite{hall2024towards}.
\item CLIP improved upon the baselines, and in particular, coverage scales with the number of guiding samples. With higher $k$ we observe reduced marginal improvements, and we deem this to the fact that the model tended to converge towards generating an average representation of the object. We hypothesize this is because CLIP's pre-training is ``concept-centric'' (aligning images to general text concepts), whereas DINOv2's is ``instance-centric'' due to its self-supervised training, making it better at preserving the unique features of a specific reference image. 
\end{itemize}

We report these results in \Cref{tab:feature_ablation}. This analysis confirms that DINOv2 is the most effective choice for our method, with CLIP being an alternative.

\begin{table}[!ht]
    \caption{Comparison of feature extractors across different $k$ values. DINOv2 shows strong improvements in $F_1$, precision, and coverage compared to \ldmonefive and CLIP, while also achieving lower FID.}
    \centering
    \small
    \setlength{\tabcolsep}{4pt}  %
    \resizebox{.9\textwidth}{!}{
    \begin{tabular}{l c S[table-format=1.4, detect-weight] S[table-format=1.4, detect-weight] S[table-format=1.4, detect-weight] S[table-format=1.4, detect-weight] S[table-format=1.4, detect-weight] S[table-format=3.2, detect-weight] S[table-format=2.2, detect-weight]}
    {\textbf{Feature Extractor}} & \textbf{$k$} & {\textbf{$\mathbf{F_1 (P, C)}$} $\uparrow$} & {\textbf{Precision} $\uparrow$} & {\textbf{Coverage} $\uparrow$} & {\textbf{Density} $\uparrow$} & {\textbf{Recall} $\uparrow$} & {\textbf{FDD} $\downarrow$} & {\textbf{FID} $\downarrow$} \\\midrule
    \ldmonefive $\omega = 1.0$ & --  & 0.2334 & 0.4363 & 0.1593 & 0.1731 & \textbf{0.6135} & 684.81 & 35.27 \\
    \rowcolor{gray!10}\ldmonefive $\omega = 2.0$ & --  & 0.3277 & 0.5433 & 0.2346 & 0.2647 & 0.4859 & 524.59 & 24.04 \\
    \ldmonefive $\omega = 7.5$ & --  & 0.2960 & 0.6222 & 0.1942 & 0.3629 & 0.2025 & 693.31 & 42.60 \\
    \rowcolor{gray!10}DINOv2 & 2  & 0.4242 & 0.6354 & 0.3184 & 0.4027 & 0.4614 & 410.84 & 19.73 \\
    DINOv2 & 4  & 0.5313 & 0.8296 & 0.3907 & 0.8933 & 0.1501 & 368.27 & 19.08 \\
    \rowcolor{gray!10}DINOv2 & 8  & 0.6251 & 0.9010 & 0.4785 & 1.3575 & 0.0708 & 353.80 & 18.81 \\
    DINOv2 & 16 & \textbf{0.7527} & \textbf{0.9461} & \textbf{0.6250} & \textbf{2.0309} & 0.0547 & \textbf{323.11} & \textbf{18.44} \\
    \rowcolor{gray!10}CLIP   & 2  & 0.3888 & 0.6367 & 0.2798 & 0.3987 & 0.3604 & 446.77 & 19.66 \\
    CLIP   & 4  & 0.4063 & 0.5951 & 0.3084 & 0.3501 & 0.4827 & 421.20 & 19.01 \\
    \rowcolor{gray!10}CLIP   & 8  & 0.4163 & 0.5974 & 0.3195 & 0.3567 & 0.5030 & 416.75 & 18.25 \\
    CLIP   & 16 & 0.4088 & 0.5757 & 0.3169 & 0.3180 & 0.5735 & 415.02 & 18.22 \\
    \end{tabular}}
    \label{tab:feature_ablation}
\end{table}

%% file: tables/strength_ablation_sd15_in1k.tex
\begin{table}[!ht]
    \caption{Chamfer guidance strength $\gamma$ ablation on ImageNet-1k using \ldmonefive. Our Chamfer guidance obtains the best coverage results when using a high ($\gamma=0.07$) strength, but obtains the most balanced image quality (FID) when using a milder amount ($\gamma=0.05$).}
    \centering
    \small
    \setlength{\tabcolsep}{4pt}  %
    \resizebox{.8\textwidth}{!}{
    \begin{tabular}{S[table-format=1.2] c S[table-format=1.3, detect-weight] S[table-format=1.3, detect-weight] S[table-format=1.3, detect-weight] S[table-format=1.3, detect-weight] S[table-format=1.3, detect-weight] S[table-format=3.3, detect-weight] S[table-format=2.3, detect-weight]}
    {$\mathbf{\gamma}$} & \textbf{$k$} & {\textbf{$\mathbf{F_1}$ (P, C)} $\uparrow$} & {\textbf{Precision} $\uparrow$} & {\textbf{Coverage} $\uparrow$} & {\textbf{Density} $\uparrow$} & {\textbf{Recall} $\uparrow$} & {\textbf{FDD} $\downarrow$} & {\textbf{FID} $\downarrow$} \\\midrule
    0.02 & 2 & 0.837 & 0.896 & 0.786 & 0.910 & \textbf{0.702} & 145.390 & 9.268 \\
    \rowcolor{gray!10}0.02 & 32 & 0.872 & 0.912 & 0.835 & 0.994 & 0.699 & 138.917 & 9.301 \\
    0.05 & 2 & 0.881 & 0.932 & 0.835 & 1.051 & 0.637 & 124.191 & \textbf{8.840} \\
    \rowcolor{gray!10}0.05 & 32 & 0.914 & 0.946 & 0.884 & 1.162 & 0.650 & 114.847 & 8.906 \\
    0.07 & 2 & 0.886 & 0.947 & 0.833 & 1.108 & 0.480 & 156.179 & 13.670 \\
    \rowcolor{gray!10}0.07 & 32 & \textbf{0.931} & \textbf{0.950} & \textbf{0.912} & \textbf{1.213} & 0.649 & \textbf{113.301} & 8.935 \\
    \end{tabular}}
    \label{tab:str_ablation_sd15_in1k}
    \end{table}

%% file: sec/appendices/obj_additional_results.tex
\section{Additional object-centric results}

In \Cref{tab:ldm_performance_comparison_cads_apg_imagenet1k} we show more results of base \ldmonefive and \ldmthreefive sampling, and additional set of parameters for CADS~\cite{sadat2023cads} and APG~\cite{sadat2024eliminatingoversaturationartifactshigh}. These results extend those presented in~\Cref{tab:quantitative_combined}, and show how these reference-free approaches cannot significantly increase the fidelity and diversity of the samples, and that applying APG on a small $\omega$ brings moderate improvement over the best base sampling, for both \ldmonefive and \ldmthreefive.

\input{tables/apg_cads_imagenet}

%% file: tables/apg_cads_imagenet.tex
\begin{table}[!ht]
\caption{Comparison of different LDM configurations with APG and CADS guidance on ImageNet-1k.}
\centering
\small
\setlength{\tabcolsep}{4pt}  %
{ 
\begin{tabular}{
  l
  cc
  S[table-format=1.1]
  S[table-format=1.3]
  S[table-format=1.3]
  S[table-format=1.3]
  S[table-format=1.3]
  S[table-format=3.1]
  S[table-format=2.1]
}
{\textbf{Method}} & 
{\textbf{APG}} & 
{\textbf{CADS}} & 
{$\mathbf{\omega}$} & 
{\textbf{$\mathbf{F_1 (P, C)} \uparrow$}} & 
{\textbf{Precision}$\uparrow$} & 
{\textbf{Coverage}$\uparrow$} & 
{\textbf{Density}$\uparrow$} & 
{\textbf{FDD}$\downarrow$} & 
{\textbf{FID}$\downarrow$} \\
\midrule
\midrule
\multicolumn{10}{l}{\textbf{LDM 1.5}} \\
\midrule
\rowcolor{blue!5}
\textit{LDM 1.5} & --- & --- & 1.0 & 0.507 & 0.723 & 0.391 & 0.551 & 431.2 & 31.3 \\
\rowcolor{blue!5}
\textit{LDM 1.5} & --- & --- & 2.0 & 0.673 & 0.802 & 0.580 & 0.648 & 226.2 & \textbf{10.8} \\
\rowcolor{blue!5}
\textit{LDM 1.5} & --- & --- & 7.5 & 0.709 & \textbf{0.862} & 0.603 & \textbf{0.775} & 248.7 & 16.1 \\
\rowcolor{gray!10}
LDM 1.5 & \checkmark & \xmark & 2.0 & 0.677 & 0.809 & 0.582 & 0.644 & 226.0 & \textbf{10.8} \\
\rowcolor{gray!10}
LDM 1.5 & \checkmark & \xmark & 4.0 & \textbf{0.723} & 0.855 & \textbf{0.626} & 0.752 & \textbf{217.9} & 13.4 \\
LDM 1.5 & \checkmark & \xmark & 7.5 & 0.713 & 0.853 & 0.612 & 0.768 & 247.9 & 16.2 \\
\rowcolor{gray!10}
LDM 1.5 & \checkmark & \xmark & 10.0 & 0.707 & 0.858 & 0.601 & 0.766 & 261.2 & 17.0 \\
LDM 1.5 & \xmark & \checkmark & 1.0 & 0.487 & 0.713 & 0.370 & 0.526 & 431.4 & 31.2 \\
\rowcolor{gray!10}
LDM 1.5 & \xmark & \checkmark & 2.0 & 0.676 & 0.806 & 0.582 & 0.641 & 226.5 & 10.9 \\
LDM 1.5 & \xmark & \checkmark & 4.0 & 0.718 & 0.850 & 0.621 & 0.743 & 218.0 & 13.4 \\
\rowcolor{gray!10}
LDM 1.5 & \xmark & \checkmark & 7.5 & 0.715 & 0.856 & 0.615 & 0.771 & 247.0 & 16.1 \\
LDM 1.5 & \xmark & \checkmark & 10.0 & 0.707 & 0.857 & 0.601 & 0.770 & 260.8 & 16.9 \\
\midrule
\multicolumn{10}{l}{\textbf{LDM 3.5}} \\
\midrule
\rowcolor{blue!5}
\textit{LDM 3.5} & --- & --- & 1.0 & 0.599 & 0.752 & 0.498 & 0.560 & 314.7 & 17.3 \\
\rowcolor{blue!5}
\textit{LDM 3.5} & --- & --- & 2.0 & \textbf{0.727} & 0.872 & 0.623 & 0.797 & \textbf{231.9} & 15.7 \\
\rowcolor{blue!5}
\textit{LDM 3.5} & --- & --- & 7.5 & 0.690 & \textbf{0.878} & 0.568 & 0.829 & 329.6 & 24.4 \\
LDM 3.5 & \checkmark & \xmark & 2.0 & 0.723 & 0.856 & \textbf{0.625} & 0.783 & 237.6 & 15.4 \\
\rowcolor{gray!10}
LDM 3.5 & \checkmark & \xmark & 4.0 & 0.722 & 0.877 & 0.613 & \textbf{0.832} & 287.4 & 21.4 \\
LDM 3.5 & \checkmark & \xmark & 7.5 & 0.695 & 0.874 & 0.576 & 0.820 & 334.7 & 23.7 \\
\rowcolor{gray!10}
LDM 3.5 & \checkmark & \xmark & 10.0 & 0.681 & 0.870 & 0.559 & 0.798 & 349.8 & 23.5 \\
LDM 3.5 & \xmark & \checkmark & 1.0 & 0.582 & 0.751 & 0.475 & 0.529 & 346.6 & 19.9 \\
\rowcolor{gray!10}
LDM 3.5 & \xmark & \checkmark & 2.0 & 0.717 & 0.851 & 0.620 & 0.749 & 238.1 & \textbf{15.1} \\
LDM 3.5 & \xmark & \checkmark & 4.0 & 0.712 & 0.870 & 0.603 & 0.807 & 282.3 & 20.9 \\
\rowcolor{gray!10}
LDM 3.5 & \xmark & \checkmark & 7.5 & 0.688 & 0.868 & 0.570 & 0.799 & 328.7 & 23.2 \\
LDM 3.5 & \xmark & \checkmark & 10.0 & 0.679 & 0.866 & 0.559 & 0.778 & 345.6 & 23.1 \\
\end{tabular}
}
\label{tab:ldm_performance_comparison_cads_apg_imagenet1k}
\end{table}

%% file: sec/appendices/dp_additional_results.tex
\section{Additional classification results}

We complement our downstream utility analysis by evaluating the downstream utility of images generated using our Chamfer Guidance. 
We follow the ``static'' ImageNet-100 setup introduced in~\cite{askari25dp}, and generate a dataset of $130,000$ synthetic images using our Chamfer Guidance, using $k=32$ real exemplar images from the training set. 
Each synthetic image is generated with a simple prompt \texttt{(class name)}. 
We report the accuracy of a ViT-B~\citep{dosovitskiy2021an} classifier trained on this synthetic data and tested on real validation data. 
In \Cref{tab:choice_of_gens} we compare to the base performance of several LDMs, and  observe  substantial gains when using our Chamfer Guidance. 
Interestingly, the initial performance gap of 6.5 points between \ldmonefive and \ldmthreefive (presumably because reduced sample diversity in \ldmthreefive) is reduced to 2.4 points when using Chamfer Guidance.
These results confirm the ability of Chamfer Guidance to improve downstream utility of synthetic data.

\begin{table}[ht]
\vspace{-1em}
\centering
\caption{Validation accuracy on real images of classifiers trained with synthetic data for ImagetNet-100 classes. All models are trained for 50k iterations.}\label{tab:choice_of_gens}
{\small
\resizebox{1.0\textwidth}{!}{

\begin{tabular}{lccc|cc|cc}
 & \textbf{\ldmonefour} & \textbf{\ldmtwoone} & \textbf{\ldmxl} & \textbf{\ldmonefive} & \textbf{\ldmonefive + Chamfer} & \textbf{\ldmthreefive} & \textbf{\ldmthreefive + Chamfer} \\ \midrule
\rowcolor{gray!10}\textbf{Real Val. Acc.} & 59.06 & 55.92 & 52.8 & 59.24 & \textbf{67.82} & 52.72 & \textbf{65.42} \\
\end{tabular}
}
}
\end{table}

%% file: sec/appendices/licenses.tex
\section{Licenses}

We report the licenses for datasets and models used in~\Cref{tab:assets}.

\begin{table}
    \centering
    \caption{Links and licenses for the datasets and pre-trained models we use in our study.}
    {\scriptsize
    \begin{tabular}{lll}
        Name & Link & License \\
        \midrule
        ImageNet & \url{https://www.image-net.org} & Apache License 2.0\\
        GeoDE & \url{https://geodiverse-data-collection.cs.princeton.edu/} &  CC BY 4.0\\
        DollarStreet & \url{https://mlcommons.org/datasets/dollar-street/} & CC-BY-SA 4.0\\
        \midrule
        DINOv2 & \url{https://github.com/facebookresearch/dinov2} & Apache License 2.0\\ 
        CLIP & \url{https://github.com/openai/CLIP} & MIT License \\
        Llama2-70b & \url{https://huggingface.co/meta-llama/Llama-2-70b-chat} &  Llama 2 Community License Agreement \\
        \midrule
        \ldmonefive & \url{https://huggingface.co/ruwnayml/stable-diffusion-v1-5} &  CreativeML Open RAIL-M \\
        \ldmthreefive & \url{https://huggingface.co/stabilityai/stable-diffusion-3.5-medium} & Stability AI Community License \\
        \midrule
        dgm-eval & \url{https://github.com/layer6ai-labs/dgm-eval} & MIT License\\
    \end{tabular} 
    }
    \label{tab:assets}
\end{table}

%% file: sec/appendices/qualitatives.tex
\section{Additional qualitative results}

\begin{figure}
    \centering
    \includegraphics[width=1.0\linewidth]{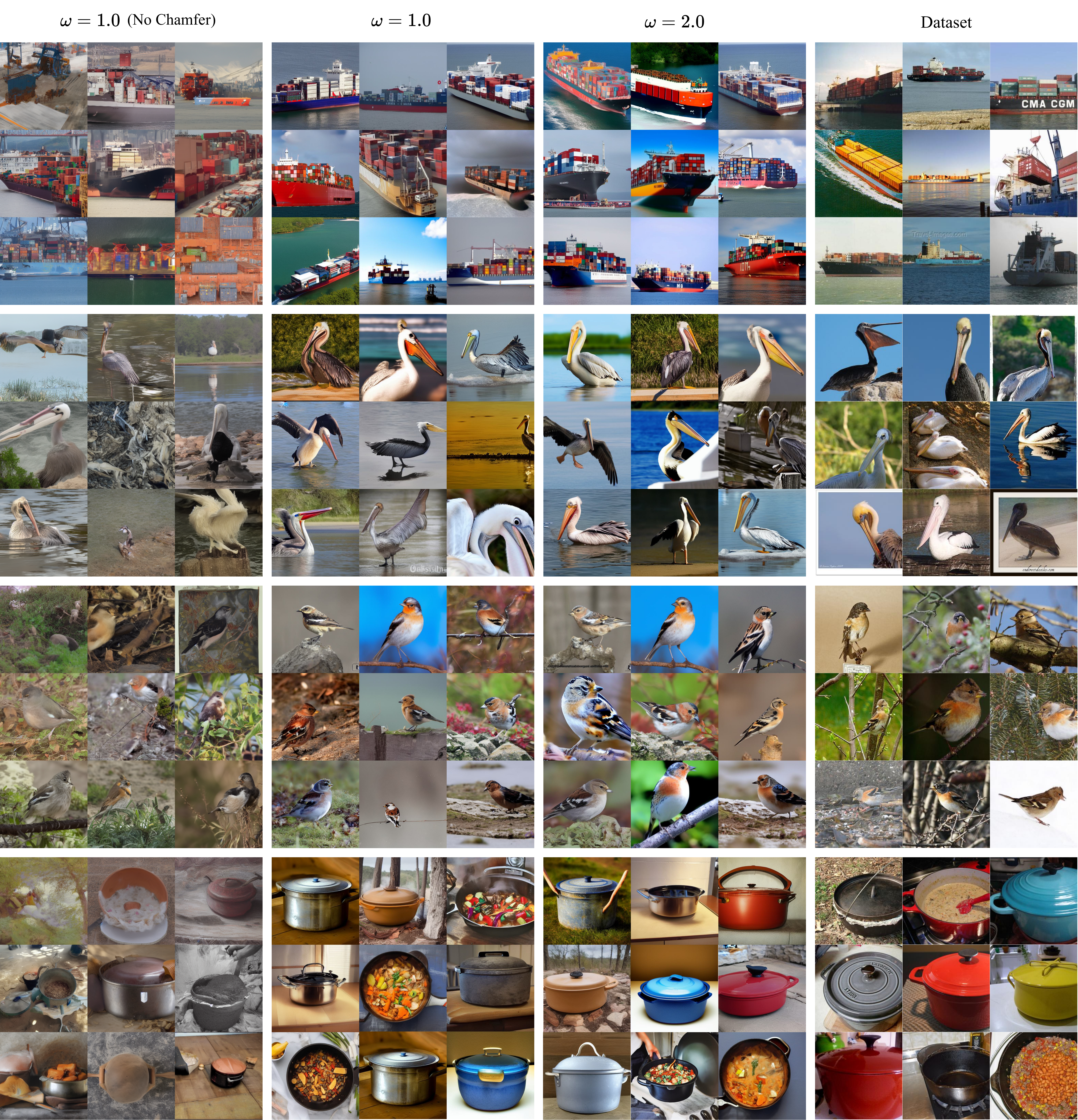}
    \caption{\ldmonefive generations on ImageNet-1k with different $\omega$ values. $k=32$, $\gamma=0.07$ for our Chamfer guidance. The classes are from top to bottom: \texttt{container ship}, \texttt{pelican}, \texttt{brambling}, and \texttt{dutch oven}.}
    \label{fig:ldm15_omega_comparison}
\end{figure}

\Cref{fig:ldm15_omega_comparison} shows \ldmonefive with $\omega=1.0$ without and with our Chamfer guidance, and $\omega=2.0$ with our Chamfer guidance. \ldmonefive does not have a ``strong'' conditional only model ($\omega=1.0$), as quantitatively highlighted in~\Cref{tab:ldm_performance_comparison_cads_apg_imagenet1k}, which exhibits poor coverage and density, and a high FID. Qualitatively, we observe bad-looking images that often lack the correct subject, or present a distorted one. Our Chamfer guidance, even when applied only the conditional model, can effectively recover the correct texture, shape, and proportions of the subject. For \ldmonefive, the best results are obtained by applying our Chamfer guidance on top of a low $\omega$ value, \eg $2.0$.

\begin{figure}
    \centering
    \includegraphics[width=1.0\linewidth]{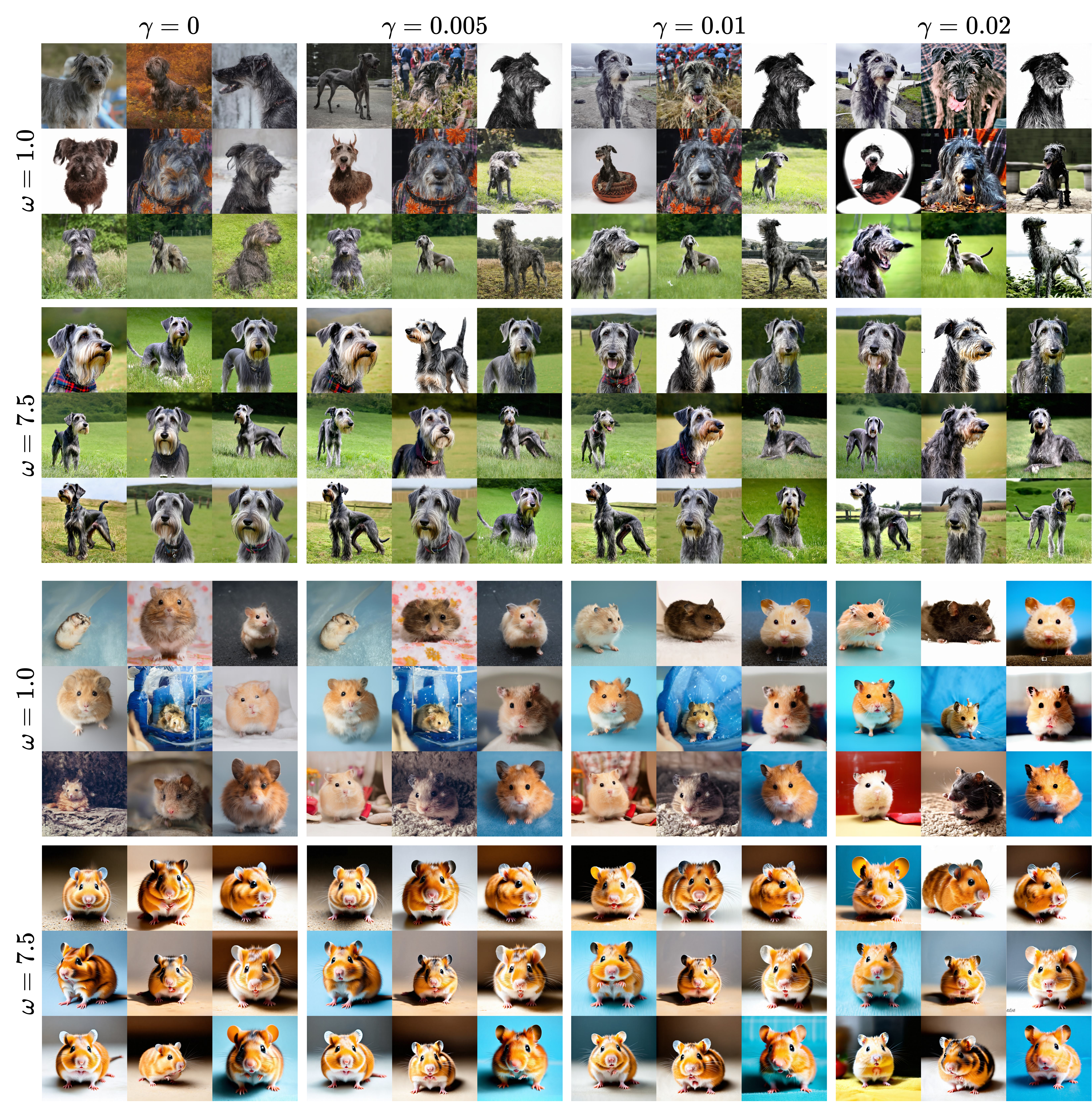}
    \caption{\ldmthreefive on ImageNet-1k with different $\omega$ and $gamma$ values, $k=32$. The classes are from top to bottom: \texttt{Irish wolfhound} and \texttt{hamster}}
    \label{fig:ldm35_omega_gamma_comparison}
\end{figure}

\Cref{fig:ldm35_omega_gamma_comparison} shows that \ldmthreefive without CFG, \ie $\omega=1.0$ produces poor quality images, often with distorted or wrong subjects. We can appreciate that Chamfer distance, applied with moderate strength $\gamma$, can effectively steer the generation towards the right subject, increasing the quality and diversity of the generation. When using a high CFG strength $\omega$ \ie $7.5$, steering the generation with our Chamfer guidance requires a higher strength $\gamma$ before becoming effective. Our Chamfer guidance helps to recover naturalness of the image by reducing oversaturation, and to increase the variety in the backgrounds.

\begin{figure}
    \centering
    \includegraphics[width=0.8\linewidth]{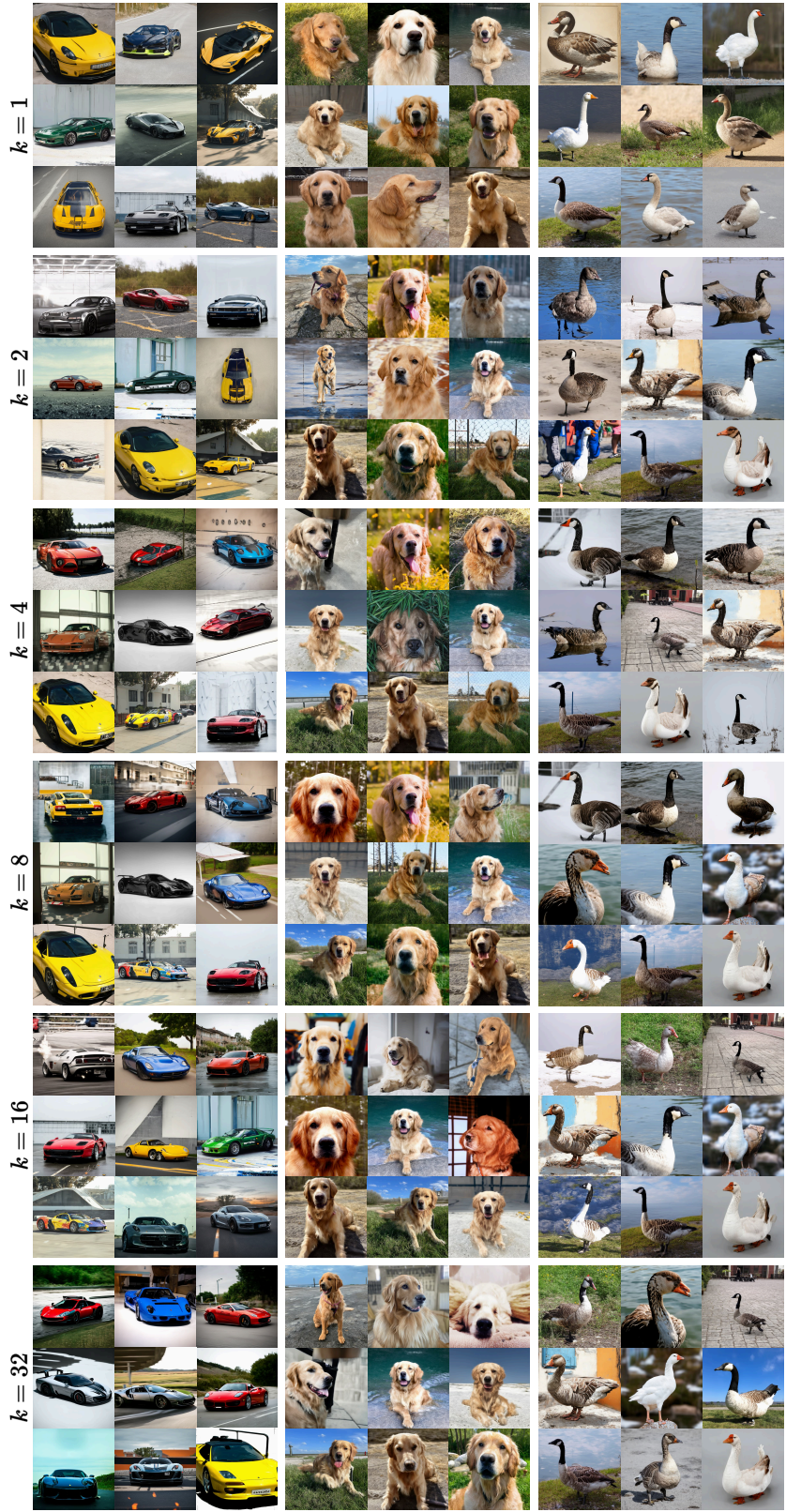}
    \caption{\ldmthreefive on ImageNet-1k with different $k$ values. $\omega$ and $\gamma$ are picked based on the best $F_1$ score. The classes from left to right are: \texttt{sports car}, \texttt{golden retriever}, and \texttt{goose}.}
    \label{fig:ldm35_k_comparison}
\end{figure}

\Cref{fig:ldm35_k_comparison} qualitatively shows the effect of increasing the number of exemplar images $k$ when using our Chamfer guidance. We can see a greater diversity in both subjects and background when using a higher number for $k$, in agreement with the quantitative results presented in~\Cref{fig:scaling_imagenet_1k}.

In \Cref{fig:qualitatives_geo} we show the comparison of the geographic diversity benchmark, in particular the GeoDE dataset. We show how base generations using \ldmonefive introduce stereotypical elements when using the geographical indicator, \eg dilapidated cars, or rusty pans. c-VSG~\cite{hemmat2024improving} partially mitigates the issue, but it comes at the expense of quality, generating oversaturated images that do not resemble the dataset. Our Chamfer guidance effectively mitigates both issues, by generating natural-looking images without stereotypical elements, due to the guidance of real exemplars.

\input{figures/cvsg_qualitatives}

%% file: figures/cvsg_qualitatives.tex
\begin{figure}[htbp]
    \centering
    
    \begin{subfigure}{1.\textwidth}
        \centering
        \includegraphics[width=1.0\textwidth]{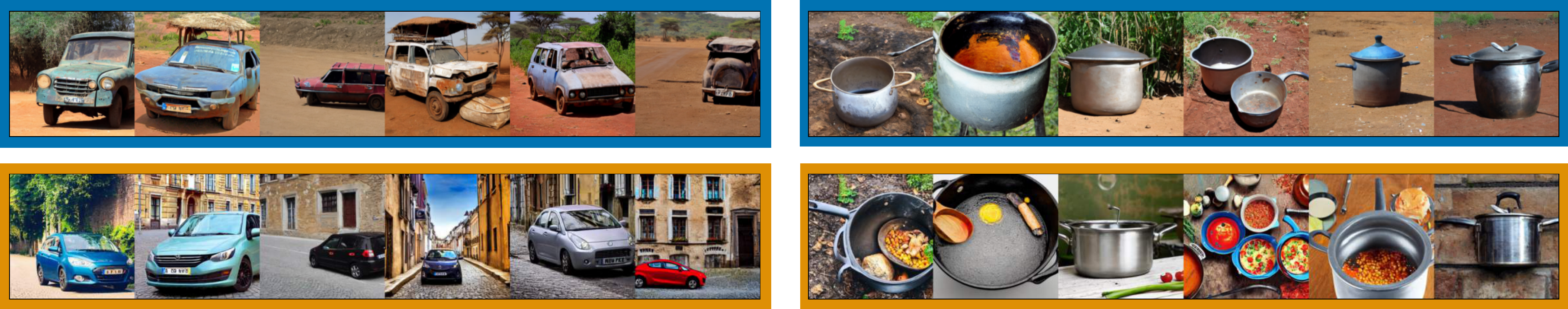}
        \caption{\texttt{\{object\}} in \texttt{\{region\}} baseline.}
        \label{fig:base_geo}
    \end{subfigure}
    
    \begin{subfigure}{1.\textwidth}
        \centering
        \includegraphics[width=1.0\textwidth]{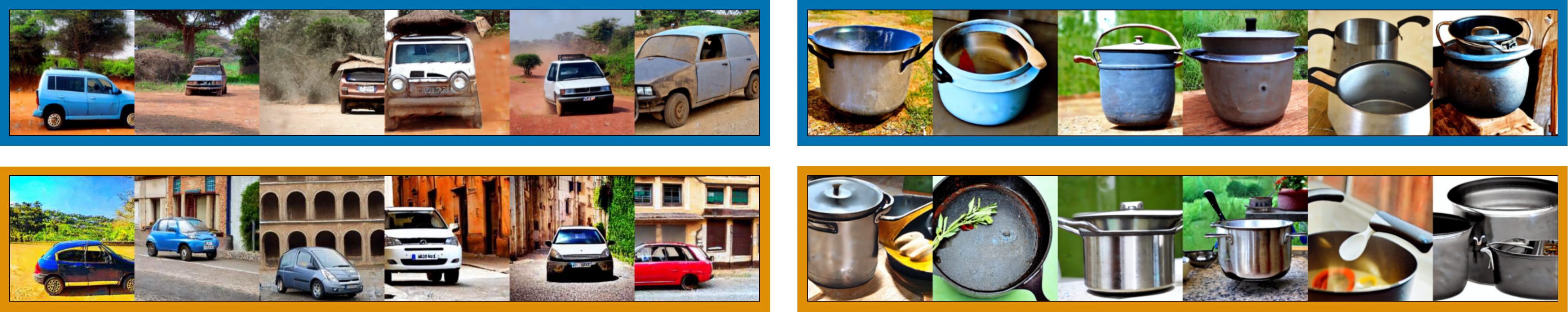}
        \caption{c-VSG~\cite{hemmat2024improving} generations.}
        \label{fig:cvsg_geo}
    \end{subfigure}
    
    \begin{subfigure}{1.\textwidth}
        \centering
        \includegraphics[width=1.0\textwidth]{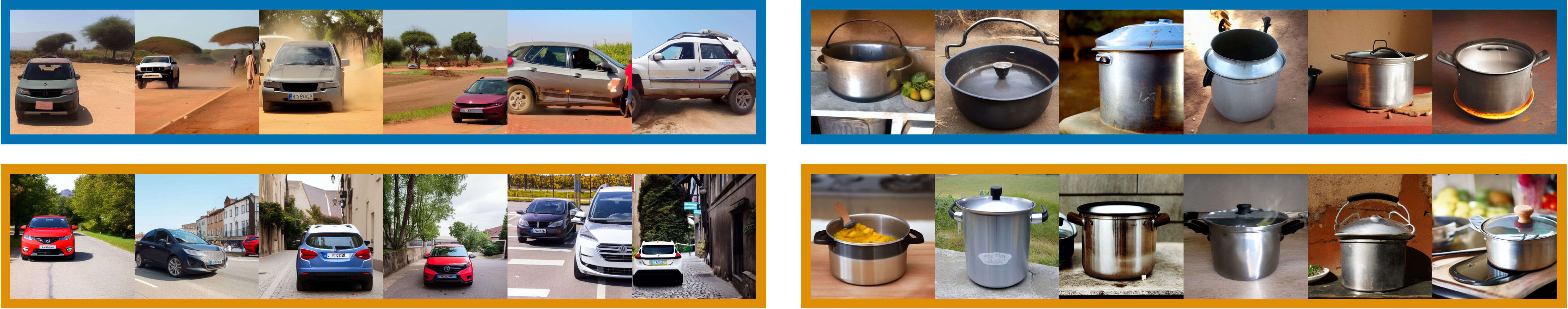}
        \caption{Our chamfer guidance.}
        \label{fig:chamfer_geo}
    \end{subfigure}

    \begin{subfigure}{1.\textwidth}
        \centering
        \includegraphics[width=1.0\textwidth]{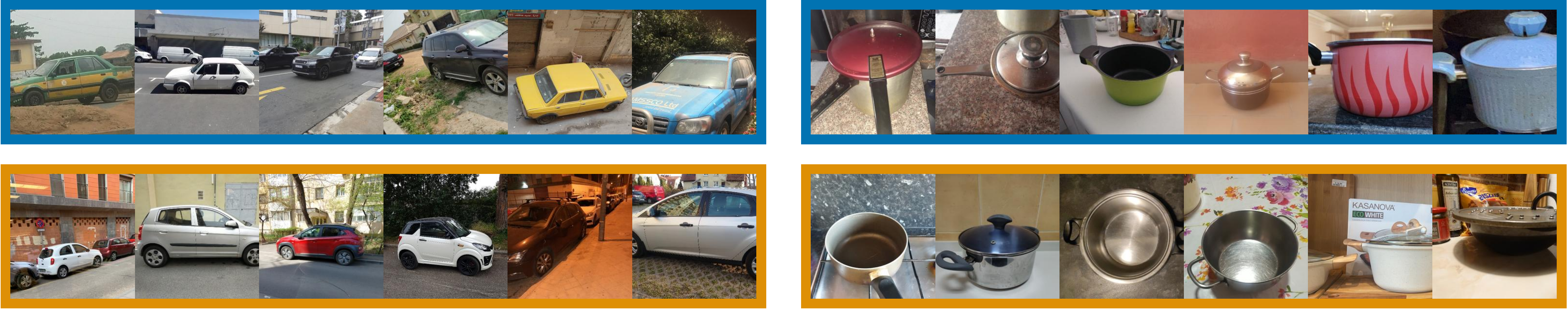}
        \caption{Samples from the GeoDE dataset.}
        \label{fig:dataset_geo}
    \end{subfigure}
    
    \caption{ Generated images and examples from the GeoDE dataset of cars (left) and cooking pots (right) using \ldmonefive~on GeoDE. Colors indicate images in \textcolor{RoyalBlue}{Africa} and \textcolor{orange}{Europe}. Our Chamfer Guidance exhibits better-looking images with less saturated colors, increased subject quality and diversity in the backgrounds.
    }
    \vspace{-2em}
    \label{fig:qualitatives_geo}
\end{figure}